\pdfoutput=1

\documentclass[11pt]{article}

\usepackage[final]{acl}

\usepackage{times}
\usepackage{latexsym}

\usepackage[T1]{fontenc}

\usepackage[utf8]{inputenc}

\usepackage{microtype}

\usepackage{inconsolata}

\usepackage{graphicx}

\usepackage{tabularx}
\usepackage{booktabs}
\usepackage{amsmath}
\usepackage{threeparttable}
\usepackage{multirow}
\usepackage{colortbl}
\usepackage{hhline}
\usepackage{xcolor}
\usepackage{subcaption}
\usepackage{caption}
\usepackage{tabularx}
\usepackage{multirow}
\usepackage{hyperref}
\usepackage{fontawesome}

\title{Lookahead Q-Cache: Achieving More Consistent KV Cache Eviction \\ via Pseudo Query}

\author{%
  Yixuan Wang\footnotemark[2]\quad{Shiyu Ji}\footnotemark[2]\quad{Yijun Liu}\quad{Yuzhuang Xu} \\
  \textbf{Yang Xu}\quad\textbf{Qingfu Zhu}\quad\textbf{Wanxiang Che}\footnotemark[1] \\
  Research Center for Social Computing and Interactive Robotics, \\
  Harbin Institute of Technology, China \\
  \texttt{\{yixuanwang,car\}@ir.hit.edu.cn} \\
  \faGithubAlt~\url{https://github.com/noforit/Lookahead_Q-Cache}
}

\begin{document}
\maketitle

\renewcommand{\thefootnote}{\fnsymbol{footnote}}
\footnotetext[2]{Equal contribution.}
\footnotetext[1]{Corresponding author.}
\renewcommand{\thefootnote}{\arabic{footnote}}

\begin{abstract}
Large language models (LLMs) rely on key-value cache (KV cache) to accelerate decoding by reducing redundant computations.
However, the KV cache memory usage grows substantially with longer text sequences, posing challenges for efficient deployment.
Existing KV cache eviction methods prune tokens using prefilling-stage attention scores, causing inconsistency with actual inference queries, especially under tight memory budgets.
In this paper, we propose Lookahead Q-Cache (LAQ), a novel eviction framework that generates low-cost pseudo lookahead queries to better approximate the true decoding-stage queries.
By using these lookahead queries as the observation window for importance estimation, LAQ achieves more consistent and accurate KV cache eviction aligned with real inference scenarios.
Experimental results on LongBench and Needle-in-a-Haystack benchmarks show that LAQ outperforms existing methods across various budget levels,
achieving a 1 $\sim$ 4 point improvement on LongBench under limited cache budget.
Moreover, LAQ is complementary to existing approaches and can be flexibly combined to yield further improvements.
\end{abstract}

\section{Introduction}
Large language models (LLMs) have demonstrated strong capabilities in long-sequence text modeling tasks \citep{liu2025comprehensive,liu2025survey}
such as code generation \citep{guo2024deepseek,hui2024qwen2}, document summarization \citep{liu2024sumsurvey}, and mathematical reasoning \citep{chen2025towards}.
To improve inference efficiency during the decoding stage,
LLMs leverage key-value cache (KV-Cache) to reduce redundant computations and accelerate inference.
However, increasing text lengths lead to a substantial rise in KV cache memory usage,
introducing considerable obstacles \citep{shi2024keep,li2024survey,yuan2024kv} to efficient model deployment.
Recent work \citep{liu2024kivi,sun2024you} has focused on lightweighting the KV cache in long-context scenarios.

\begin{figure}[t]
  \includegraphics[width=\linewidth]{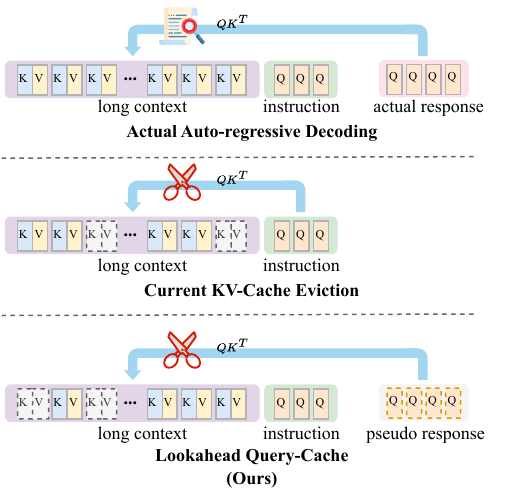}
  \caption{Illustration of the differences between the proposed Lookahead approach and existing methods.
Pseudo queries are introduced to probe the importance of cached keys and values.}
  \label{fig:intro}
\end{figure}

As a straightforward and effective compression method,
KV cache eviction has attracted widespread attention \citep{ge2023model,liu2023scissorhands,tang2024quest} from researchers.
The method improves decoding efficiency by performing token-level pruning on the prefilling-stage KV cache.
Following the observations of \citet{liu2023scissorhands,zhang2023h2o},
the majority of studies adopt cumulative attention scores as the criterion for token pruning.
Compared to direct accumulation, SnapKV \citep{li2024snapkv} achieves more accurate KV cache importance estimation by leveraging an observation window.
In addition, some studies \citep{cai2024pyramidkv,feng2024ada} aim to allocate finer-grained budgets across layers or attention heads to achieve higher compression rates.

Although existing methods have partially alleviated the KV cache overhead in long-context scenarios,
several challenges remain to be addressed.
One key issue is the \textbf{inconsistency between compression and inference}.
Under a constrained budget, KV cache eviction seeks to maintain the key-value pairs most likely to be accessed by future queries during decoding.
However, as shown in Figure \ref{fig:intro}, current approaches rely on prefilling-stage queries to approximate those in the decoding stage
when selecting key-value pairs for KV cache retention.
This inconsistency significantly reduces the accuracy of cache selection under a low budget,
thereby degrading the performance of eviction algorithms.

In this paper, we observe that employing the prefix of the generated response as the observation window leads to more consistent KV cache selection.
Notably, this phenomenon is insensitive to the quanlity of the generated response:
even incorrect answers are often able to recall the key cache entries required for generating the correct one.
Motivated by these findings,
we propose a KV cache eviction framework, \textbf{LookAhead Q-Cache (LAQ)}, which aligns more closely with the actual inference situation.
Specifically, unlike prior works that select observation windows from the input text to compute the importance of KV cache,
our method generates lookahead queries in a low-cost manner that are more consistent with the actual response queries.
Once the lookahead queries are cached,
we use them as the observation window for importance estimation, and then proceed with standard KV cache eviction.
Experimental results on LongBench and Needle-in-a-Haystack tasks demonstrate that LAQ
consistently outperforms existing methods across various budget settings.
By leveraging Q-Cache, which better aligns with inference scenarios,
the proposed method even outperforms some dynamically budgeted approaches under low-budget settings.
Furthermore, experiments show that the proposed method can be integrated with existing methods to
yield orthogonal improvements.

The main contributions of this paper are as follows:
\begin{itemize}
  \item We propose the Lookahead Q-Cache,
a novel framework that mitigates the inconsistency issue during both compression and inference phases
by leveraging pre-generated low-quality pseudo queries.
  \item The proposed method offers orthogonal improvements when combined with existing approaches,
and its flexible configuration enables broad applicability.
  \item Experiments across diverse benchmarks indicate that the proposed method significantly outperforms existing strategies,
consistently achieving performance improvements of 1 $\sim$ 4 percentage points on LongBench.
\end{itemize}

\section{Related Works}
\subsection{KV Cache Eviction}
KV Cache eviction aims to prune redundant KV cache entries from the prefill stage at the token level,
improving decoding speed and alleviating memory usage without compromising decoding performance.
Previous work primarily builds on \citeposs{liu2023scissorhands,zhang2023h2o} findings,
utilizing the attention mechanism as an inherent evaluation criterion to assess the importance of KVCache entries and guide their eviction.
However, using global attention scores often lacks specificity.
\citet{li2024snapkv} discover that the suffix of the input window exhibits behavior more consistent with the generation phase.
By maintaining a local window to guide KVCache eviction, lower performance degradation is achieved.

Due to conflicts between obtaining attention scores and existing acceleration techniques (such as FlashAttention \citep{dao2022flashattention,dao2023flashattention}),
some methods focus on identifying alternative importance evaluation metrics.
\citet{devoto2024simple} identify a correlation between the norm of the key values and their importance,
and propose an L2 norm-based KV cache eviction method that is compatible with existing acceleration frameworks.
Building on this, \citet{guo2024attention} also explore the norm of the value vectors
and achieve more precise KV Cache eviction by combining the attention mechanism with the L2 norm of Value caches.
Despite alleviating performance degradation, existing KV Cache eviction strategies continue to
face the inconsistency problem in compressed inference illustrated in Figure \ref{fig:intro},
thereby constraining the upper bound of eviction performance.

\subsection{KV Cache Compression}
In addition to direct eviction, some studies \citep{wan2024d2o} focus on merging similar KV cache entries to improve efficiency.
To preserve as much information from the prefill stage as possible, some methods \citep{bolya2022token,zhang2024cam} build upon KV cache eviction
and further explore token-level cache fusion, leading to improved task performance after eviction.
Going further, with appropriate processing and training \citep{sun2024you,liu2024minicache,zuhri2024mlkv,ma2024compressing}, KV cache entries across layers can be merged and compressed to reduce memory consumption, offering additional efficiency gains.

Moreover, dimensionality reduction of the KV cache  \cite{shazeer2019fast} serves as an effective approach to improve efficiency.
The most straightforward compression method is to reduce the precision bit-width \citep{he2024zipcache} of the KV Cache.
\citet{liu2024kivi} propose separate quantization methods based on channels and tokens to address the distinct distribution characteristics of the key and value,
achieving nearly lossless 2-bit quantization.
Additionally, some methods employ low-rank techniques for compression.
GQA \citep{ainslie2023gqa} achieves significant compression by sharing keys and values across multiple groups of queries,
and has been widely adopted in various applications \citep{grattafiori2024llama}.
Multi-Head Latent Attention (MLA) \citep{liu2024deepseek} of DeepSeek significantly reduces KV cache size through low-rank compression and decoupled RoPE,
achieving efficient inference while maintaining model performance.

\section{Observasions}
\label{sec:obser}

\begin{figure}[t]
  \centering
  \includegraphics[width=\linewidth]{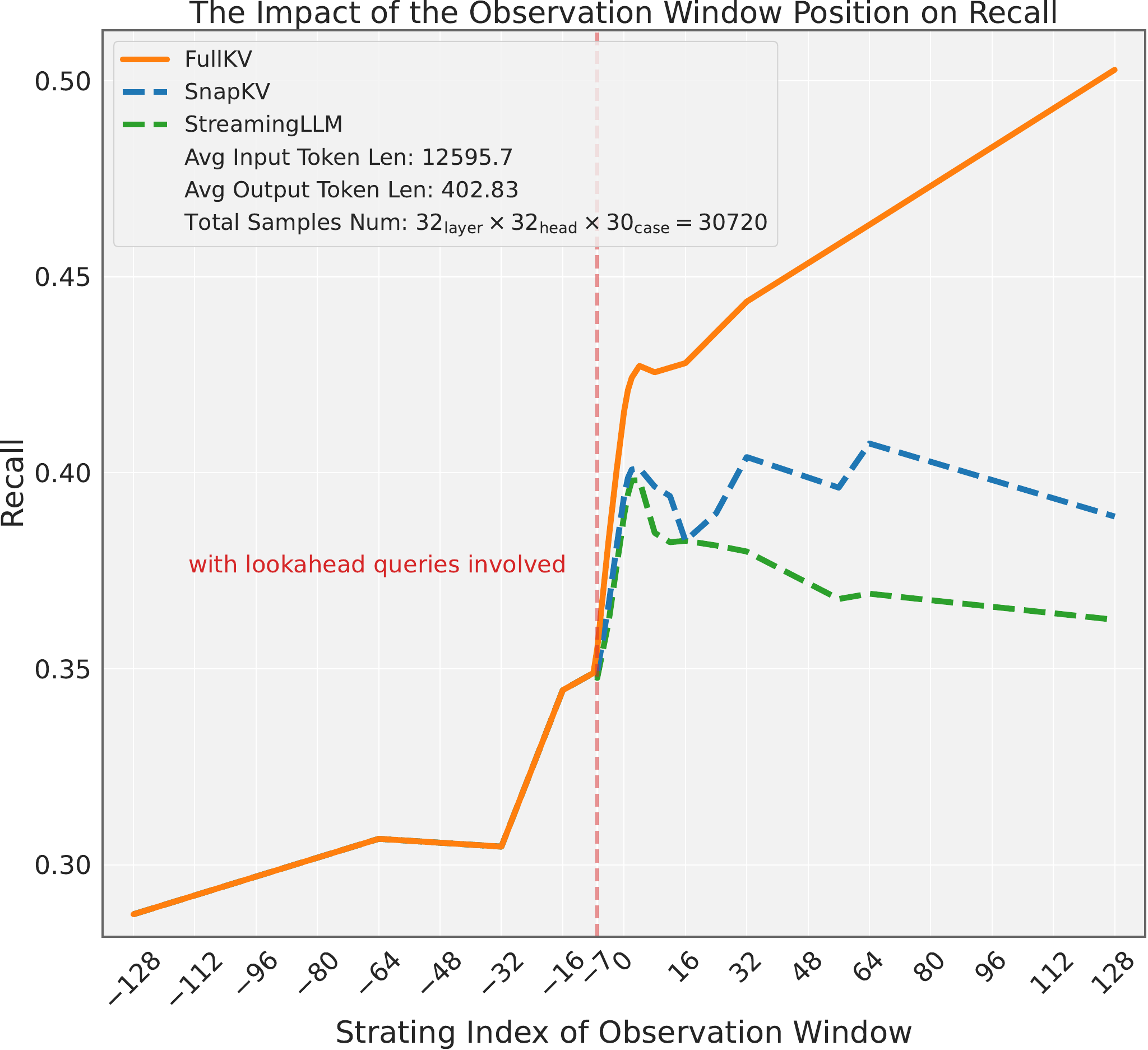}
  \caption{An illustration of the average recall rate with different starting index of the observation window,
where 0 indicates the position of the first generated token.
The observation window has a fixed length of 8, and the budget is set to 1024.}
  \label{fig:obs}
\end{figure}

To investigate the inconsistency between the compression and inference stages,
we first evaluate the recall rate of the selected KV cache using fix-length observation windows at different positions.
Specifically, the position includes not only the input portion but also the output tokens generated during the inference stage.
The recall rate of the selected KV cache is defined
as the proportion of indices selected by the observation window that overlap with those selected by all response tokens.
The final recall rate $Recall_{W}$ can be calculated as follows:
\begin{equation}
  M_{gold} = \mathrm{ArgSort}(\sum_{i \in R}{q_{i}}K^T)[\mathrm{:Budget}]
\end{equation}
\begin{equation}
  M_{pred} = \mathrm{ArgSort}(\sum_{i \in W}{q_{i}}K^T)[\mathrm{:Budget}]
\end{equation}
\begin{equation}
  \label{eq:recall}
  Recall_{W} = \frac{\vert M_{pred} \cap M_{gold} \vert}{\vert M_{gold} \vert}
\end{equation}
As shown in Figure \ref{fig:obs}, we conduct evaluations on the GovReport dataset \citep{huang2021efficient} for the summarization task.
By analyzing the figure, we obtain several insightful findings.

\paragraph{(1) Response prefix queries significantly boost recall over instruction.}
Similar to the conclusion of \citet{li2024snapkv},
we find that local windows near the end of the input achieve higher recall rates compared to other input windows.
However, once tokens from the output portion appear in the observation window (as indicated by the red vertical line in the figure),
the recall rate of the window experiences a sudden increase.
Although the recall rate gradually increases as the window shifts forward,
just 1-2 tokens from the generation phase are sufficient to bring a substantial improvement.
This phenomenon indicates a significant discrepancy between input and output queries
and suggests that existing eviction methods still have substantial room for improvement.

\paragraph{(2) Queries from low-quality responses remain effective.}
Although output queries are effective, they are not accessible during the actual inference stage.
To address this, we further evaluate the recall rate of observation windows composed of queries from low-quality responses generated under common KV-Cache eviction strategies.
As shown by the dashed lines in Figure \ref{fig:obs}, the recall rate is lower than that achieved with the full KV cache,
yet a noticeable jump in recall can still be observed with these low-quality queries.
This suggests that the selection of KV cache may not be highly sensitive to response correctness.
Actually, cache patterns attended by incorrect responses remain more consistent
with those of correct responses than traditional input-based selections.

Based on the above insights, if we can obtain the query corresponding to the response prefix as a observation window,
the consistency between the compression and inference stages can be improved,
leading to substantial gains for existing KV-Cache eviction strategies.
Notably, although the golden query is not accessible during inference,
some pseudo queries obtained through eviction strategies can still achieve strong performance.
These pieces of evidence motivate the design of a two-stage KV-Cache eviction algorithm,
which enables more precise cache eviction by leveraging pseudo response-prefix windows with more consistent distributions.

\section{Lookahead Q-Cache}

\subsection{Overall Workflow}
\begin{figure}[t]
  \includegraphics[width=\linewidth]{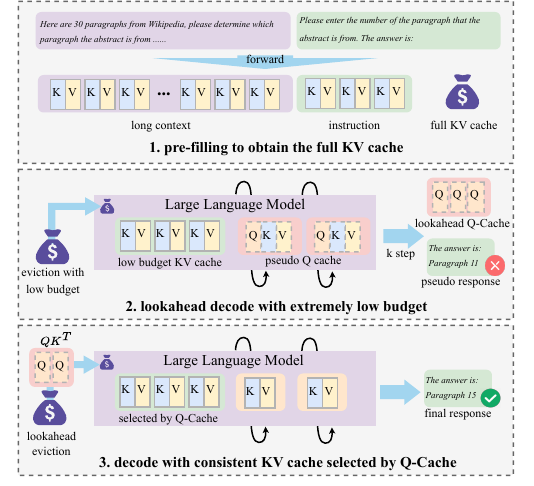}
  \caption{Main workflow of the proposed Lookahead Q-Cache.
  Queries from incorrect answers can still retrieve KV cache entries aligned with the actual outputs.}
  \label{fig:main}
\end{figure}

Given the outcomes in Section \ref{sec:obser},
the proposed LookAhead Q-Cache (LAQ) aims to cache queries from some low-quality responses in advance as the observation window to achieve KV-Cache eviction more consistent with the inference stage.
Our main workflow, as illustrated in Figure \ref{fig:main}, consists of three stages: the prefilling stage, the lookahead stage (Section \ref{sec:la}), and the eviction-based decoding stage (Section \ref{sec:re}).

\subsection{Lookahead Query Construction}
\label{sec:la}
After obtaining the KV cache during the pre-filling stage,
we first employ certain KV cache selection strategies to construct lookahead queries.
To balance efficiency, we adopt KV cache eviction methods (such as StreamingLLM \citep{xiao2023efficient} and snapKV \citep{li2024snapkv}) with a low budget
to generate a specified number of response tokens at a lower cost.
Considering that the primary bottleneck for model generation in long-text scenarios lies in the pre-filling stage, a few additional decoding steps introduce minimal impact on the overall latency.
We further analyze the latency introduced by the Lookahead Q-Cache in Section \ref{sec:latency}.

Unlike standard generation, the lookahead generation process requires retaining the Query hidden states as a Q-Cache for use in subsequent observation windows.
However, since we decode only up to 8 steps ahead, the additional memory overhead introduced by the Q-Cache is negligible in the context of long sequences.

\subsection{KV-Cache Re-eviction}
\label{sec:re}
Given the previously obtained Q-Cache $Q$,
we use it as an observation window to perform a second-round eviction of the KV-Cache.
The retained KV-Cache indices under budget $B$ can be formulated as:
\begin{equation}
  \label{eq:la_index}
  M_{ahead} = \mathrm{ArgSort}(\sum_{q_i \in Q}{q_{i}}K^T)[:B]
\end{equation}
After obtaining the KV cache ranking under the lookahead observation window,
we can perform cache re-eviction based on the given budget,
followed by standard autoregressive decoding for generation.
Specifically, the pseudo Query cache can be combined with the queries from the local window
obtained during the prefilling stage to form a unified observation window for generation:
\begin{equation}
  \label{eq:la++_index}
  M_{ahead} = \mathrm{ArgSort}(\sum_{q_i \in W \cup  Q}{q_{i}}K^T)[:B]
\end{equation}
where $W$ represents the local context window, similar to that employed in SnapKV \citep{li2024snapkv}.
We denote this integrated method as LAQ++.

\begin{table*}[!t]

\fontsize{20}{24}\selectfont
\setlength{\tabcolsep}{5pt}
\centering
\caption{Performance comparison of different methods across various LLMs on LongBench.}\label{tab:longbench}
\begin{threeparttable}
\scalebox{0.3}{
\begin{tabular}{llccccccccccccccccc}
\specialrule{1pt}{0pt}{2pt}
&\multirow{4}{*}{\textbf{~~~LLMs}} & \multicolumn{3}{c}{\textbf{Single-Document QA}} & \multicolumn{3}{c}{\textbf{Multi-Document QA}}& \multicolumn{3}{c}{\textbf{Summarization}}& \multicolumn{3}{c}{\textbf{Few-shot Learning}}& \multicolumn{2}{c}{\textbf{Synthetic}} & \multicolumn{2}{c}{\textbf{Code}}&\multirow{4}{*}{\textbf{~~~Avg.}} \\
\cmidrule(lr){3-5}\cmidrule(lr){6-8}\cmidrule(lr){9-11}\cmidrule(lr){12-14}\cmidrule(lr){15-16}\cmidrule(lr){17-18}
&& \rotatebox[origin=c]{30}{\textbf{NrtvQA}} & \rotatebox[origin=c]{30}{\textbf{Qasper}} & \rotatebox[origin=c]{30}{\textbf{MF-en}} & \rotatebox[origin=c]{30}{\textbf{HotpotQA}} & \rotatebox[origin=c]{30}{\textbf{2WikiMQA}} & \rotatebox[origin=c]{30}{\textbf{Musique}} & \rotatebox[origin=c]{30}{\textbf{GovReport}} & \rotatebox[origin=c]{30}{\textbf{QMSum}} & \rotatebox[origin=c]{30}{\textbf{MultiNews}} & \rotatebox[origin=c]{30}{\textbf{TREC}} & \rotatebox[origin=c]{30}{\textbf{TriviaQA}} & \rotatebox[origin=c]{30}{\textbf{SAMSum}} & \rotatebox[origin=c]{30}{\textbf{PCount}} & \rotatebox[origin=c]{30}{\textbf{PRe}} & \rotatebox[origin=c]{30}{\textbf{Lcc}} & \rotatebox[origin=c]{30}{\textbf{RB-P}} \\

\specialrule{1pt}{2pt}{2pt}

\multirow{19}{*}{\rotatebox[origin=c]{90}{\fontsize{18}{100}\selectfont \textbf{Mistral-7B-v0.2-Instruct}}}

&~~~Full KV & 26.90 & 33.14 & 49.26 & 42.77 & 27.35 & 18.77 & 32.95 & 24.21 & 27.13 & 71.00 & 86.23 & 42.72 & 2.75 & 86.98 & 57.12 & 54.51 & 42.74 \\
\cline{2-19}

& \multicolumn{18}{c}{\cellcolor{lightgray!25} 
  \textbf{KV Cache Size = 128}
} \\


&~~~H2O & 21.62 & 21.34 & 38.61 & 30.46 & 20.38 & 12.20 & 20.59 & 22.51 & 22.03 & 39.00 & 82.33 & 40.64 & 3.06 & \textbf{80.56} & 51.17 & 48.46 & 34.69 \\

&~~~SnapKV & 19.71 & 21.13 & 42.75 & 36.45 & 22.36 & 15.76 & 19.05 & 21.81 & 21.36 & 47.50 & 84.15 & 40.29 & 2.41 & 68.26 & 52.26 & 48.75 & 35.25 \\

&~~~PyramidKV & 21.98 & 22.78 & 43.78 & 32.30 & 22.31 & 15.81 & 20.41 & 21.82 & 21.23 & 66.00 & 83.51 & 39.83 & 2.99 & 65.81 & 51.61 & 46.42 & 36.16 \\


&~~~\textbf{LAQ} & 24.94 & \textbf{27.77} & 45.43 & 40.35 & 25.25 & 17.91 & \textbf{22.06} & 22.68 & \textbf{22.50} & 69.50 & \textbf{86.36} & 41.16 & 1.49 & 76.85 & 53.31 & \textbf{51.02} & 39.29 \\


&~~~\textbf{LAQ++} &\textbf{25.62} & 27.21 & \textbf{46.16} & \textbf{40.60} & \textbf{25.93} & \textbf{18.44} & 21.60 & \textbf{23.07} & 22.42 & \textbf{70.00} & 86.18 & \textbf{42.03} & \textbf{3.51} & 74.81 & \textbf{54.68} & 50.92 & \textbf{39.57} \\

\cline{2-19}
& \multicolumn{18}{c}{\cellcolor{lightgray!25} 
  \textbf{KV Cache Size = 256}
} \\


&~~~H2O & 21.42 & 23.04 & 42.60 & 30.75 & 22.42 & 13.82 & 22.35 & 22.54 & 23.12 & 40.50 & 83.78 & 40.73 & 3.51 & 85.85 & 53.18 & 49.95 & 36.22 \\

&~~~SnapKV & 22.44 & 24.07 & 48.01 & 38.66 & 22.66 & 15.59 & 21.83 & 23.23 & 22.94 & 61.50 & 85.45 & 41.32 & 3.13 & 85.79 & 55.11 & 51.73 & 38.97 \\

&~~~PyramidKV & 21.69 & 25.18 & 47.61 & 38.77 & 26.12 & 15.23 & 22.52 & 22.52 & 22.59 & 68.00 & 84.27 & 42.10 & 3.43 & 76.60 & 53.08 & 48.40 & 38.63 \\


&~~~\textbf{LAQ} & 24.68 & \textbf{29.25} & 48.00 & 40.56 & 26.01 & 18.24 & \textbf{24.04} & 22.96 & \textbf{23.80} & 70.00 & 85.81 & 42.52 & 2.01 & 82.34 & 54.96 & \textbf{53.00} & 40.51 \\


&~~~\textbf{LAQ++} & \textbf{25.23} & 29.16 & \textbf{49.24} & \textbf{41.70} & \textbf{26.85} & \textbf{18.62} & 23.73 & \textbf{23.69} & 23.38 & \textbf{70.50} & \textbf{86.24} & \textbf{42.54} & \textbf{3.36} & \textbf{86.11} & \textbf{55.59} & 52.49 & \textbf{41.15} \\
\cline{2-19}
& \multicolumn{18}{c}{\cellcolor{lightgray!25} 
  \textbf{KV Cache Size = 512}
} \\


&~~~H2O & 21.72 & 25.54 & 44.72 & 32.39 & 23.16 & 14.75 & 23.61 & 23.03 & 24.58 & 42.00 & 85.22 & 41.61 & \textbf{3.42} & 86.45 & 54.82 & 51.11 & 37.38 \\

&~~~SnapKV & 24.14 & 28.11 & 48.78 & 39.49 & 25.09 & 17.40 & 23.67 & 23.18 & 24.60 & 67.00 & 85.88 & 41.39 & 2.78 & \textbf{86.56} & 56.61 & 53.47 & 40.51 \\

&~~~PyramidKV & 22.99 & 28.74 & 48.45 & 39.73 & 25.74 & 16.58 & 24.48 & 23.40 & 24.52 & 70.00 & 85.99 & 42.40 & 3.32 & 81.63 & 55.93 & 52.38 & 40.39 \\


&~~~\textbf{LAQ} & 24.65 & \textbf{31.21} & 49.15 & 39.90 & \textbf{27.18} & 18.38 & 25.55 & 23.91 & 24.87 & \textbf{71.00} & 86.33 & 42.14 & 1.87 & 86.41 & \textbf{56.84} & 53.08 & 41.40 \\


&~~~\textbf{LAQ++} & \textbf{25.49} & 30.92 & \textbf{49.72} & \textbf{41.50} & 26.84 & \textbf{19.20} & \textbf{25.67} & \textbf{24.04} & \textbf{25.31} & \textbf{71.00} & \textbf{86.43} & \textbf{43.14} & 2.90 & 85.27 & 56.80 & \textbf{53.54} & \textbf{41.70} \\

\midrule

\multirow{19}{*}{\rotatebox[origin=c]{90}{\fontsize{18}{100}\selectfont \textbf{Llama3.1-8B-Instruct}}}

&~~~Full KV & 31.85 & 15.55 & 28.17 & 29.93 & 22.98 & 18.20 & 34.39 & 23.79 & 27.12 & 72.50 & 92.14 & 43.66 & 8.37 & 97.59 & 65.05 & 54.78 & 41.63 \\
\cline{2-19}

& \multicolumn{18}{c}{\cellcolor{lightgray!25} 
  \textbf{KV Cache Size = 128}
} \\


&~~~H2O & 25.06 & 7.09 & 18.58 & 17.86 & 19.88 & 9.14 & 22.28 & 22.68 & 21.55 & 40.00 & 90.89 & 40.78 & 8.30 & 92.96 & 59.15 & 49.36 & 34.10 \\

&~~~SnapKV & 24.65 & 7.29 & 22.01 & 19.11 & 18.85 & 11.07 & 20.48 & 21.62 & 20.16 & 47.50 & 90.24 & 40.47 & \textbf{10.75} & 92.51 & 59.99 & 49.08 & 34.74 \\

&~~~PyramidKV & 24.79 & 8.29 & 20.72 & 14.86 & 13.84 & 8.90 & 22.41 & 22.76 & 21.53 & 62.00 & 90.35 & 39.23 & 9.27 & 93.51 & 58.77 & 46.46 & 34.86 \\


&~~~\textbf{LAQ} & 27.80 & \textbf{10.66} & 24.86 & 20.64 & 20.04 & 15.40 & \textbf{24.18} & 23.09 & \textbf{22.88} & \textbf{72.00} & 91.55 & \textbf{43.43} & 9.04 & \textbf{95.85} & 61.12 & 50.33 & 38.30 \\


&~~~\textbf{LAQ++} & \textbf{28.65} & 10.65 & \textbf{26.04} & \textbf{24.23} & \textbf{21.56} & \textbf{15.67} & 23.50 & \textbf{23.74} & 22.75 & \textbf{72.00} & \textbf{91.95} & 42.19 & 8.37 & 94.81 & \textbf{62.18} & \textbf{50.95} & \textbf{38.70} \\
\cline{2-19}
& \multicolumn{18}{c}{\cellcolor{lightgray!25} 
  \textbf{KV Cache Size = 256}
} \\


&~~~H2O & 26.01 & 8.42 & 19.69 & 17.28 & 18.21 & 9.91 & 23.64 & 22.89 & 23.20 & 41.50 & 91.29 & 41.60 & 8.00 & 94.31 & 60.79 & 50.31 & 34.82 \\

&~~~SnapKV & 28.05 & 9.83 & 22.71 & 21.48 & 19.36 & 10.96 & 22.86 & 22.75 & 22.98 & 58.00 & 92.28 & 40.87 & 8.10 & 95.30 & 63.64 & 51.35 & 36.91 \\

&~~~PyramidKV & 26.40 & 10.08 & 22.46 & 15.20 & 16.38 & 8.60 & 23.86 & 22.93 & 23.17 & 69.00 & 90.99 & 40.60 & \textbf{8.42} & 93.74 & 60.59 & 48.11 & 36.28 \\


&~~~\textbf{LAQ} & 28.86 & 12.40 & 26.44 & 21.80 & 20.91 & 15.77 & \textbf{25.83} & 23.30 & \textbf{24.26} & \textbf{72.50} & \textbf{93.08} & 42.57 & 7.53 & \textbf{95.80} & 63.51 & 51.09 & 39.10 \\


&~~~\textbf{LAQ++} & \textbf{30.25} & \textbf{12.43} & \textbf{26.63} & \textbf{25.77} & \textbf{22.83} & \textbf{18.45} & 25.07 & \textbf{23.67} & 23.75 & 72.00 & 91.97 & \textbf{43.17} & 6.93 & 94.59 & \textbf{64.07} & \textbf{53.38} & \textbf{39.69} \\
\cline{2-19}
& \multicolumn{18}{c}{\cellcolor{lightgray!25} 
  \textbf{KV Cache Size = 512}
} \\


&~~~H2O & 25.44 & 8.35 & 20.97 & 20.08 & 19.23 & 9.51 & 24.44 & 23.50 & 24.35 & 44.00 & 92.10 & 41.16 & 7.43 & 96.41 & 62.73 & 51.77 & 35.72 \\

&~~~SnapKV & 30.34 & 10.75 & 23.54 & 24.65 & 21.55 & 12.98 & 24.82 & 23.15 & 24.61 & 68.00 & \textbf{92.33} & 42.16 & 7.83 & 96.86 & 64.74 & \textbf{53.60} & 38.87 \\

&~~~PyramidKV & 28.38 & 11.59 & 25.02 & 20.06 & 18.80 & 10.64 & 25.73 & 24.03 & 25.01 & 70.00 & 92.22 & 41.73 & \textbf{8.47} & 96.42 & 63.44 & 51.02 & 38.29 \\


&~~~\textbf{LAQ} & \textbf{30.89} & \textbf{14.04} & 25.86 & 26.00 & 23.19 & 17.73 & \textbf{27.07} & 24.01 & \textbf{25.29} & \textbf{72.50} & 92.25 & \textbf{43.13} & 6.96 & \textbf{96.97} & 64.87 & 52.58 & 40.21 \\


&~~~\textbf{LAQ++} & 29.64 & 13.22 & \textbf{26.79} & \textbf{27.58} & \textbf{23.49} & \textbf{18.63} & 26.94 & \textbf{24.04} & 25.21 & \textbf{72.50} & 92.25 & 42.83 & 8.43 & 96.25 & \textbf{65.00} & 53.37 & \textbf{40.39} \\

\specialrule{1pt}{2pt}{0pt}
\end{tabular}
}
\end{threeparttable}\vspace{-10pt}
\end{table*}

\section{Experiments}
\subsection{Setup}
\label{sec:setup}

\paragraph{Evaluation benchmarks and model setup.}
Following prior works \citep{li2024snapkv,cai2024pyramidkv}, we evaluate the proposed method using the LongBench benchmark \citep{bai2023longbench} and the Needle-in-a-Haystack test \citep{kamradt2023needle}.
For LongBench, we adopt three KV cache budget settings: 128, 256, and 512.
To comprehensively evaluate the generalization ability of the proposed method,
we conduct experiments on Mistral-7B-v0.2-Instruct \citep{jiang2023mistral7b}, Llama3.1-8B-Instruct \citep{grattafiori2024llama}, and Qwen2.5-7B-Instruct \citep{yang2025qwen2}.
Due to space limitations, additional experimental results are presented in Appendix \ref{app:res}.

\paragraph{Selected baselines.}
To highlight the effectiveness of our proposed method,
we select three commonly used and strong KV-Cache eviction strategies as baselines:
\textbf{(1) H2O} \citep{zhang2023h2o} evaluates and evicts existing KV cache entries based on cumulative attention scores.
\textbf{(2) SnapKV} \citep{li2024snapkv} utilizes a local observation window to
achieve more accurate importance estimation and employs average pooling to retain a more coherent KV Cache.
\textbf{(3) PyramidKV} \citep{cai2024pyramidkv} leverages the characteristics of attention distribution across layers by allocating different budgets to the KV cache at shallow and deep layers,
thereby achieving lower performance degradation under low-budget settings.

\paragraph{Method config.}
In the main experiments, we employ LAQ and LAQ++ as the proposed methods for comparison.
Specifically, for LAQ, we adopt a configuration of 8 forward Q-Caches,
while for LAQ++, the setup consists of 8 local observation windows in addition to 8 forward Q-Caches.
Additionally, we employ SnapKV as the lookahead method, ensuring consistency in budget and experimental settings.
We further discuss the impact of hyperparameter choices in LAQ on performance in Sections \ref{sec:qua} and \ref{sec:len}.

\begin{figure*}[t]\footnotesize
  \centering
  \begin{subfigure}{0.48\linewidth}
    \centering
    \includegraphics[width=\linewidth]{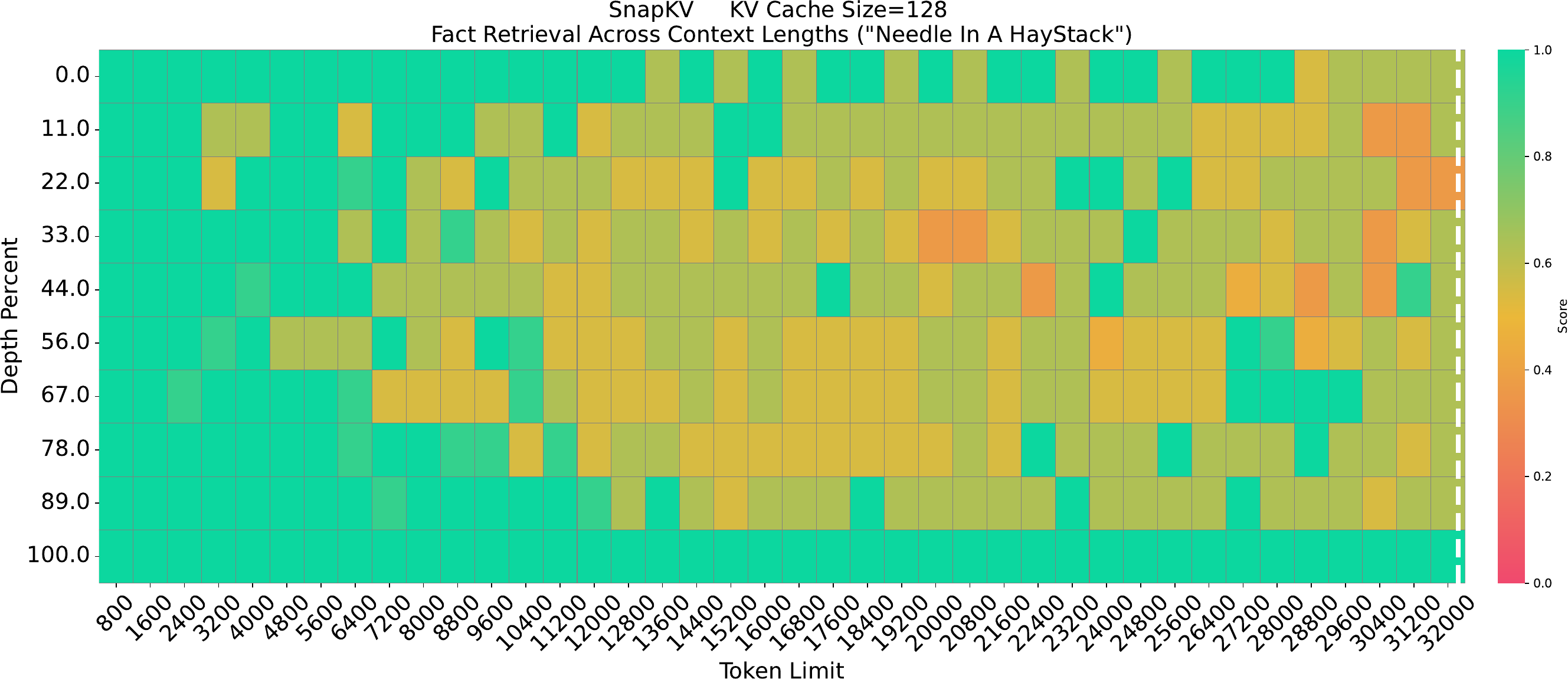}
    \caption{Mistral SnapKV (Score=76.2)}
    \label{fig:needle_snap}
  \end{subfigure}
  \hfil
  \begin{subfigure}{0.48\linewidth}
    \centering
    \includegraphics[width=\linewidth]{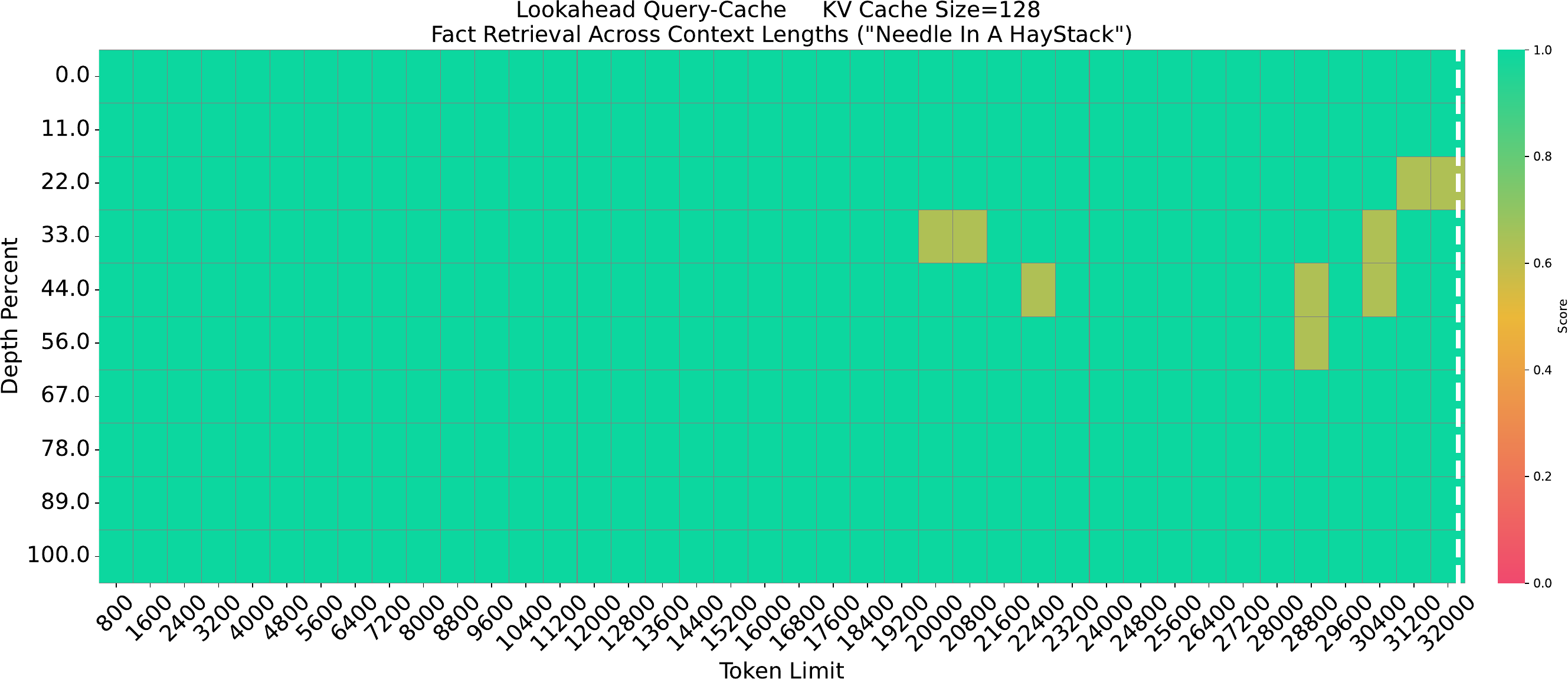}
    \caption{Mistral LAQ++ (Score=99.2)}
    \label{fig:needle_laqc}
  \end{subfigure}
  \caption{The results of SnapKV and LAQ++ on the needle-in-a-haystack test under a budget setting of 128.}
  \label{fig:needle}
\end{figure*}

\subsection{Results on LongBench.}
\paragraph{Overall.}
As shown in Table \ref{tab:longbench},
we evaluate the performance of various methods across multiple models on the LongBench benchmark.
Overall, the proposed methods achieve a significant improvement of 1$\sim$4 percentage points
over existing KV cache eviction strategies across different models and budget settings.
By leveraging a pre-fetched lookahead Q-Cache,
LAQ is able to make cache selection decisions that are more aligned with those in the actual inference stage.
Furthermore, by integrating local observation windows, LAQ++ achieves an orthogonal improvement.

\paragraph{Results across diverse output length settings.}
Moreover, beyond the overall performance gains,
LAQ also demonstrates consistently promising improvements across tasks with varying output lengths.
The construction of an 8-step lookahead Query cache enables LAQ to achieve a justified advantage on tasks characterized by shorter output lengths.
Nonetheless, it is observed that the proposed method maintains strong performance on summarization tasks where the output length exceeds 300 tokens.
This clearly demonstrates that the proposed method can partially mitigate the discrepancy of queries between the compression and inference stages.
It achieves this by incorporating pseudo-queries within the observation window.
As a result, the method attains performance improvements that are independent of output length.

\begin{table}[t]\footnotesize
  \centering
  \caption{The score on Needle-in-a-Haystack task under different KV cache budgets.}
  \label{tab:needle_results}
  \begin{tabularx}{0.92\linewidth}{c *{5}{>{\centering\arraybackslash}X}}
    \toprule
    \multirow{2}{*}{\textbf{Budget}} & \multicolumn{5}{c}{\textbf{Method}} \\
   \cmidrule(){2-6}
     & FullKV & H2O	& SnapKV	& PyraKV & LAQ++ \\
    \midrule
    \multicolumn{6}{c}{\cellcolor{lightgray!25} Mistral-7B-v0.2-Instruct} \\
     64  & 100 & 43.2 & 60.7 & 84.3 & \textbf{99.3} \\
                                   96  & 100 & 54.1 & 71.5 & 88.2 & \textbf{99.6} \\
                                   128 & 100 & 59.2 & 76.2 & 88.1 & \textbf{99.2} \\
    \midrule
    \multicolumn{6}{c}{\cellcolor{lightgray!25} Llama3.1-8B-Instruct} \\
    64  & 100 & 35.4 & 68.4 & 80.7 & \textbf{99.8} \\
                                  96  & 100 & 42.1 & 72.0 & 85.9 & \textbf{100}  \\
                                  128 & 100 & 46.1 & 73.1 & 89.9 & \textbf{100} \\
  \midrule
    \multicolumn{6}{c}{\cellcolor{lightgray!25} Qwen2.5-7B-Instruct} \\
    64  & 100 & 46.7 & 72.8 & 69.3 & \textbf{85.1} \\
                                  96  & 100 & 50.2 & 75.0 & 84.8 & \textbf{89.5} \\
                                 128 & 100 & 53.0 & 76.6 & 89.7 & \textbf{92.9} \\
    \bottomrule
  \end{tabularx}
\end{table}

\paragraph{Performance under varying budgets.}
In addition to varying models, we also conduct experiments under diverse KV cache budget settings.
Compared to baseline methods, we find that LAQ achieves more pronounced improvements under low budget settings.
Such low-budget scenarios typically impose greater demands on the recall accuracy of the KV cache within the observation window.
This result further highlights the significant gap between queries in the input and output windows,
and demonstrates that incorporating a limited number of lookahead queries can effectively mitigate this issue.
Furthermore, to ensure a fair comparison, the proposed method employs a fixed budget
and has already outperformed the dynamic-budget PyramidKV approach in low-budget scenarios.
It is conceivable that combining both approaches could yield greater improvements by enabling more fine-grained budget allocation.

\subsection{Results on Needle-in-a-Haystac Test.}
\paragraph{Overall setting.}
In addition to various tasks on LongBench, we also evaluate the proposed method on the Needle-in-a-Haystack test.
This test is designed to evaluate ability of LLMs to retrieve specific keys within ultra-long contexts.
Since FullKV already achieves strong performance with a 32K context length, in KV cache eviction scenarios, we typically conduct evaluations after evicting a given budget from the KV cache.

\begin{figure*}[t]
  \centering
  \begin{subfigure}{0.4\linewidth}
    \centering
    \includegraphics[width=\linewidth]{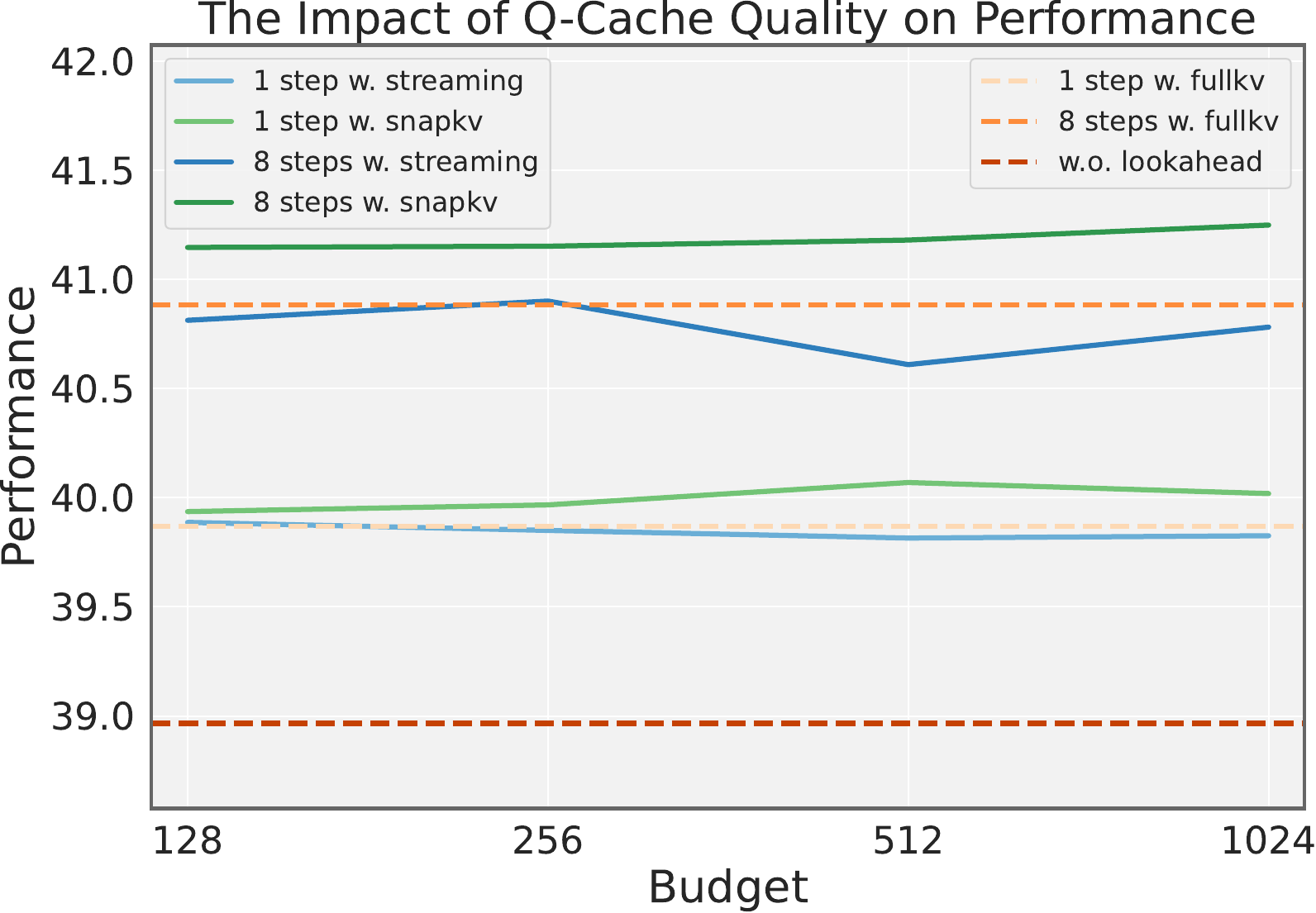}
    \caption{The impact of Q-Cache quality on performance.}
    \label{fig:quality}
  \end{subfigure}
  \hspace{0.1\textwidth}
  \begin{subfigure}{0.4\linewidth}
    \centering
    \includegraphics[width=\linewidth]{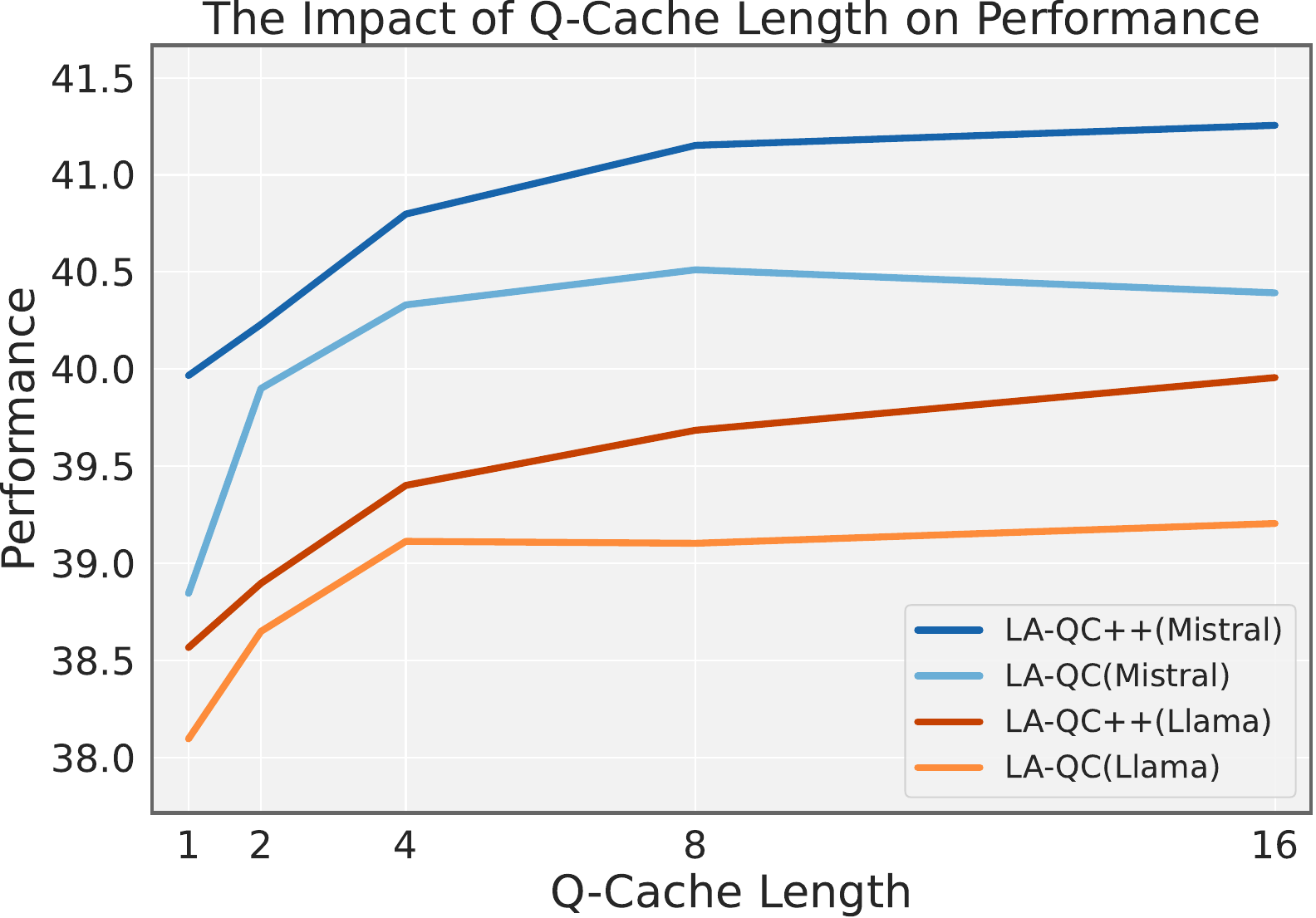}
    \caption{The impact of Q-Cache length on performance.}
    \label{fig:length}
  \end{subfigure}
  \caption{Ablation analysis of Q-Cache with respect to quality and length.}
  \label{fig:qcache}
\end{figure*}

\paragraph{Main results.}
As shown in Table \ref{tab:needle_results}, we conduct experiments on various models and budgets under a 32K context length.
With a complete KV cache available, the model demonstrates strong capability in successfully completing the retrieval task.
However, after evicting a certain portion of the KV cache, all methods experience some degree of performance degradation.
The retrieval experiment results exhibit less variability and are more intuitive compared to task performance metrics.
Although PyramidKV achieves significant advantages among baseline methods through dynamic budgeting,
LAQ++ attains near-lossless performance by leveraging a more consistent Q-Cache mechanism.
As discussed above, our proposed method can also be further enhanced by integrating dynamic budgeting.
A more intuitive example is shown in Figure \ref{fig:needle}, where compared to SnapKV, LAQ++ can recall nearly all ``needles'' from a 32K context using only a 128 budget.
Further experimental results can be found in Appendix \ref{app:needle}.

\begin{figure*}[t]
  \centering
  \begin{subfigure}{0.42\linewidth}
    \centering
    \includegraphics[width=\linewidth]{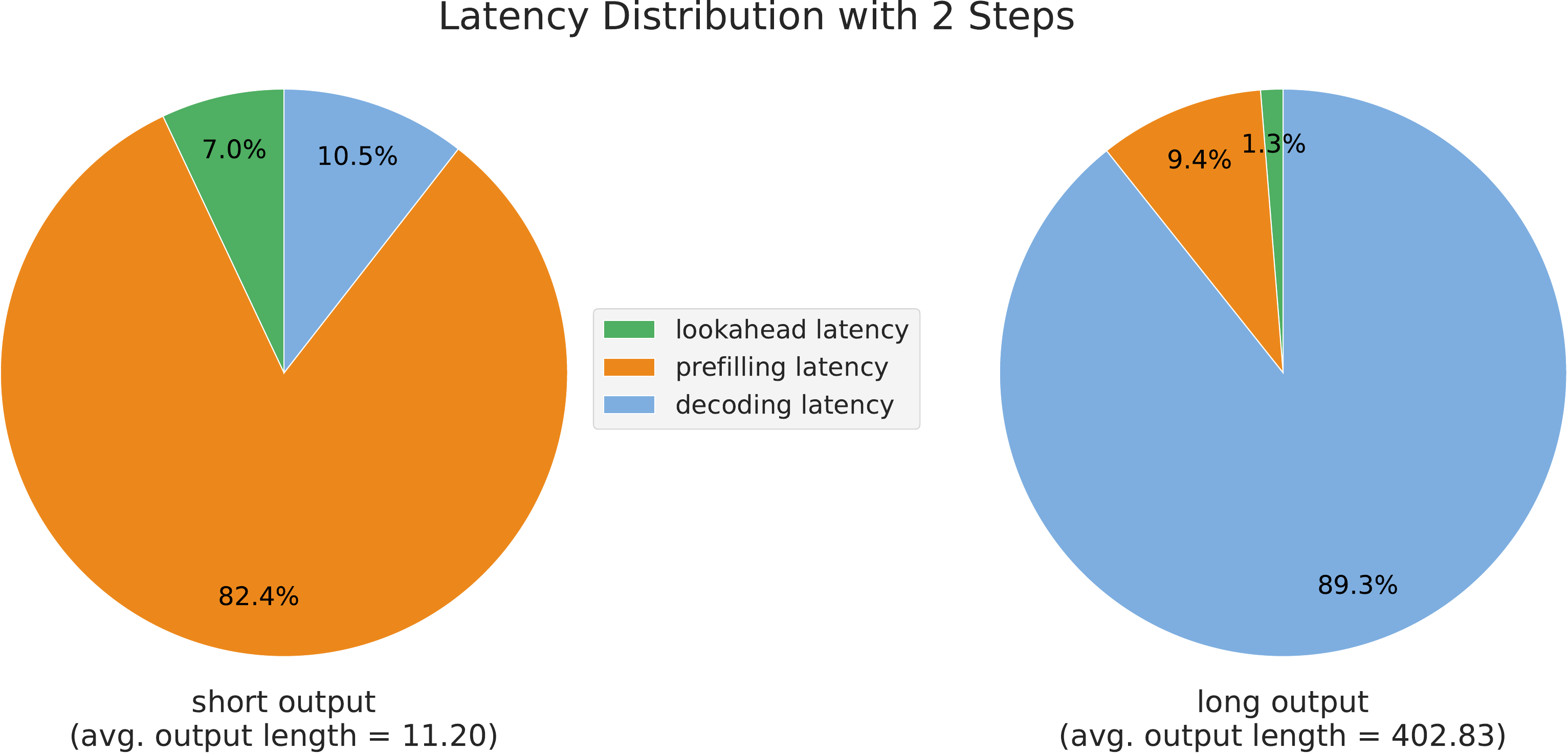}
    \caption{Lookahead Q-Cache with 2-step lookahead.}
  \end{subfigure}
  \hspace{0.1\textwidth}
  \begin{subfigure}{0.42\linewidth}
    \centering
    \includegraphics[width=\linewidth]{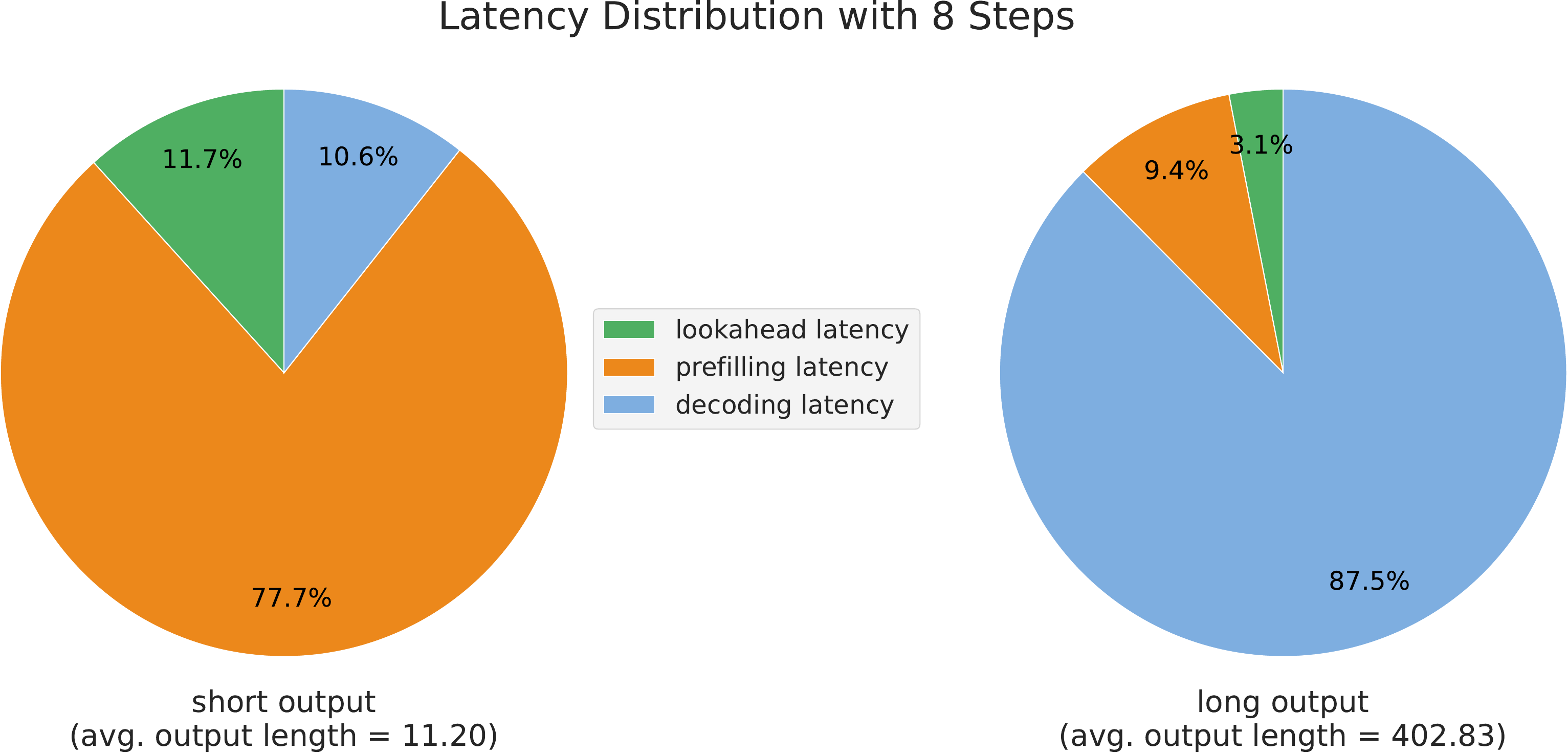}
    \caption{Lookahead Q-Cache with 8-step lookahead.}
  \end{subfigure}
  \caption{Latency breakdown under long- and short-output scenarios for 2-step and 8-step decoding,
  with green segments indicating the additional latency introduced by the proposed method.
}
  \label{fig:latency}
\end{figure*}

\begin{figure*}[t]
  \centering
  \begin{subfigure}{0.40\linewidth}
    \centering
    \includegraphics[width=\linewidth]{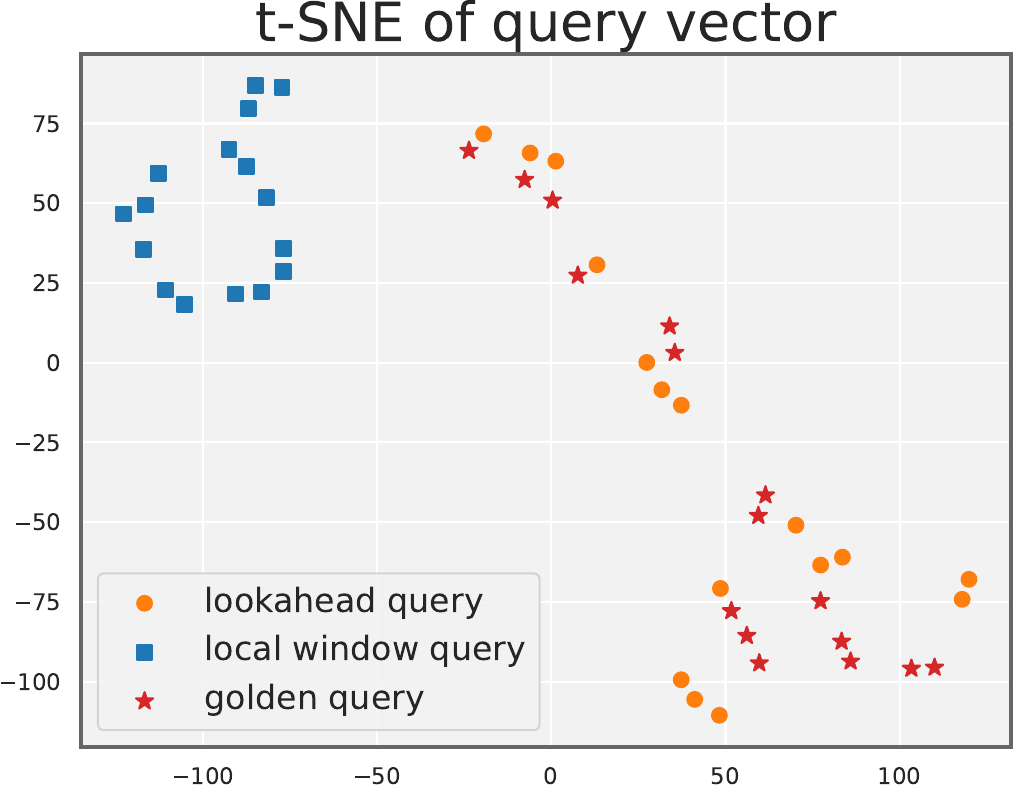}
    \caption{Visualization of query distribution after dimensionality reduction by t-SEN algorithm.}
    \label{fig:tsne}
  \end{subfigure}
  \hspace{0.15\textwidth}
  \begin{subfigure}{0.35\linewidth}
    \centering
    \includegraphics[width=\linewidth]{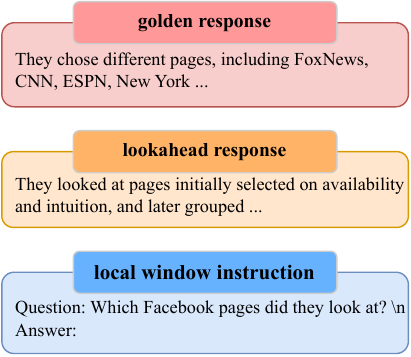}
    \caption{Text comparison of the local window instruction, lookahead response, and golden response.}
    \label{fig:context}
  \end{subfigure}
  \caption{Case study of lookahead phase.}
  \label{fig:case}
\end{figure*}

\section{Analysis}
\subsection{The Impact of Q-Cache Quality}
\label{sec:qua}
As observed in Section \ref{sec:obser}, queries from low-quality responses still achieve a high recall rate of target KV cache entries
compared to those derived from the input.
In this subsection, we quantitatively analyze the impact of Q-Cache quality on model performance.
Specifically, we employ Lookahead with different budgets and eviction strategies to obtain Q-Cache of varying quality,
and we also use the queries obtained under the Full-KV setting as an upper-bound reference.
For the configuration of the LAQ method, we conduct experiments on LongBench using two different settings:
1-step and 8-step Lookahead.

The relationship between overall performance and Q-Cache quality is shown in Figure \ref{fig:qcache}(\ref{sub@fig:quality}).
The x-axis represents the budget used in the Lookahead stage for each eviction strategy,
while the y-axis indicates the final performance on LongBench based on the KV cache entries selected by the Q-Cache.
As shown in the table,
reducing the budget has a limited impact on the performance of the proposed method.
In addition, under the same number of steps, the performance differences among different eviction strategies are also minimal.
Notably, under the 8-step LAQ setting,
the Q-Cache generated by the full KV-Cache exhibits worse task performance than that produced by SnapKV with a lower budget.
These results further demonstrate the robustness of the proposed method to the quality of the Q-Cache,
as responses generated with lower budgets can still effectively mitigate inconsistencies between the compression and inference stages.
Compared to the quality of the Q-Cache, the length has a more significant impact on the results.

\subsection{The Impact of Q-Cache Length}
\label{sec:len}
As described above, compared to the budget, the number of lookahead steps exerts a greater influence on the results.
Therefore, we further investigate the impact of different LAQ configurations combined with varying numbers of Lookahead steps on the experimental results.
As shown in Figure \ref{fig:qcache}(\ref{sub@fig:length}),
the final task performance increases with the length of the Q-Cache across all settings.
This is consistent with the recall trend observed in Figure \ref{fig:obs}.
Moreover, as the performance of low-budget eviction methods gradually degrades during long-sequence generation,
the overall effectiveness of subsequent LAQ also tends to converge.
Considering the additional latency introduced by lookahead stage,
a balanced configuration must be chosen to trade off between performance and efficiency.
In the main experiments, we set the step number hyperparameter to 8.

\subsection{Latency Analysis}
\label{sec:latency}
Despite achieving significant performance improvements,
the proposed method still incurs additional latency for lookahead operations.
To balance task performance and efficiency,
we conduct a stage-wise latency analysis of the proposed LAQ (LookAhead Q-Cache) framework.
As shown in Figure \ref{fig:latency}, we evaluate the latency distributions under both 2-step and 8-step configurations.
We evaluate on two LongBench scenarios corresponding to short-output (avg. length 11.20 tokens)
and long-output (avg. length 402.83 tokens) contexts.
The additional latency introduced by the proposed method is indicated by the green segments.

As evidenced by the results, both 2-step and 8-step configurations achieve significant performance improvements
while contributing only 2 $\sim$ 10\% to the total latency.
In short-output scenarios, the prefilling phase dominates the latency budget,
making the overhead of additional steps negligible.
For long-output scenarios, the substantial number of decoding steps renders the overhead of processing a few extra tokens marginal,
typically accounting for less than 5\% of the total runtime.

\subsection{Case Study}
To further investigate how the proposed method improves the final KV cache eviction performance through the use of the pseudo Q-Cache,
we conduct a case analysis of the queries and responses generated during the lookahead stage.
As shown in Figure \ref{fig:case},
we select a specific case to illustrate the response content and the distribution among the corresponding Q-Cache.

In Figure \ref{fig:case}(\ref{sub@fig:context}),
we present the local context window, along with the low-quality response generated in advance and the gold response generated using the full KV cache.
The three responses differ semantically in their textual content.
Due to the constraints with low budget KV cache, the lookahead response fails to provide a sufficiently accurate answer.
However, as shown in Figure \ref{fig:case}(\ref{sub@fig:tsne}),
dimensionality reduction reveals a striking consistency between the Query representations of the two responses across the three text segments.
This also validates our observation above that the gap between the input and output queries is greater than
that between the incorrect outputs and the correct outputs.
Our proposed method leverages this feature by employing a pre-generated Q-Cache
to achieve more consistent KV cache eviction.

\section{Conclusion}
In this paper, we investigate the inconsistency between the Query entries of input and output.
Based on our observational findings, we propose the Lookahead Q-Cache,
which performs KV cache eviction by using pre-generated pseudo responses as the observation window.
Experiments across multiple backbones demonstrate that
the proposed method significantly outperforms existing KV cache eviction techniques,
achieving improvements of 1 to 4 points on LongBench.

\section*{Limitations} 

Although the additional computational cost introduced by our method is minimal,
there remains room for further optimization, particularly with respect to latency.
Given the potential overlap between pseudo and golden responses,
one promising direction is to integrate the Lookahead Q-Cache mechanism with acceleration techniques
such as speculative decoding.
This combination could reduce unnecessary computations and improve real-time performance.
Building on this idea, our future work will focus on designing a more efficient KV Cache strategy that unifies these complementary approaches,
aiming to achieve consistent improvements in both task performance and inference efficiency.

\section*{Acknowledge}
We gratefully acknowledge the support of the National Natural Science Foundation of China (NSFC)
via grant 62236004, 62206078, 62441603 and 62476073.


\bibliography{custom_fix}

\appendix

\section{Detailed Latency Comparison}
\label{app:latency}
To provide a more intuitive comparison of our method with other baselines,
such as SnapKV,
we conducted additional latency analysis experiments.
These experiments were performed on a single A100-80GB GPU.
We calculated the average wall-clock time (in seconds) for each method at different stages,
with a fixed output length of 256 tokens and varying input lengths.

\begin{table}[h!]
    \centering
    \small
    \caption{Average wall-clock time (in seconds) for different methods. The output length is fixed at 256 tokens.}
    \label{tab:latency_analysis}
    \begin{tabular}{lrrr}
        \toprule
        \textbf{Method} & \textbf{Pre-filling} & \textbf{Stage 2} & \textbf{Decoding} \\
        \midrule
        \multicolumn{4}{c}{\textbf{INPUT 16K}} \\
        \midrule
        Full KV & 1.46615 & - & 12.38489 \\
        StreamingLLM & 1.49165 & - & 8.79209 \\
        H2O & 1.47285 & - & 8.62985 \\
        SnapKV & 1.48718 & - & 8.64949 \\
        LAQ++ (2) & 1.46968 & 0.12636 & 8.75256 \\
        LAQ++ (8) & 1.46952 & 0.31302 & 8.61714 \\
        \midrule
        \multicolumn{4}{c}{\textbf{INPUT 32K}} \\
        \midrule
        Full KV & 3.58076 & - & 18.63942 \\
        StreamingLLM & 3.65054 & - & 8.73857 \\
        H2O & 3.58777 & - & 8.75228 \\
        SnapKV & 3.61811 & - & 8.61096 \\
        LAQ++ (2) & 3.59406 & 0.22643 & 8.83214 \\
        LAQ++ (8) & 3.59088 & 0.41255 & 8.64124 \\
        \bottomrule
    \end{tabular}
\end{table}

As shown in the Table \ref{tab:latency_analysis},
the additional Stage 2 in our method incurs a minimal time overhead, typically less than 5\%.
This minor increase in latency is a small trade-off for the significant performance enhancements in the eviction algorithm,
particularly under low-budget conditions (see Table \ref{tab:longbench} in the main text and Tables \ref{tab:longbenchMistral} and \ref{tab:all_longbench} in the appendix).
This demonstrates the practical value of our approach.

\section{Analysis of Forward Templates}
\label{app:prompt}
To investigate the impact of different forward templates,
we conducted experiments using templates such as "The answer is" and "I have found the answer!".
These templates were designed to provide a fixed semantic guide without requiring additional decode computation.
The goal was to determine if the performance gains observed in our method, LAQ++, were due to semantic guidance
rather than the query vector information derived from lookahead.
The Table \ref{tab:forward_templates} shows the experimental results on the LongBench benchmark
with a 512 KV-Cache budget for two different base models:
Mistral-7B-v2.0-Instruct and Llama3.1-8B-Instruct.
\begin{table}[h]
    \centering
    \small
    \caption{LongBench scores for different forward templates with a 512 KV-Cache budget.}
    \label{tab:forward_templates}
    \begin{tabular}{llc}
        \toprule
        \textbf{Base Model} & \textbf{Method} & \textbf{Score} \\
        \midrule
        \multirow{7}{*}{Mistral} 
        & Full KV & 42.74 \\
        & Streaming & 31.63 \\
        & H2O & 37.38 \\
        & SnapKV & 40.51 \\
        & "The answer is" & 38.20 \\
        & "I have found the answer!" & 37.79 \\
        & LAQ++ & \textbf{41.70} \\
        \midrule
        \multirow{7}{*}{Llama} 
        & Full KV & 41.63 \\
        & Streaming & 34.61 \\
        & H2O & 35.72 \\
        & SnapKV & 38.87 \\
        & "The answer is" & 36.86 \\
        & "I have found the answer!" & 36.59 \\
        & LAQ++ &  \textbf{40.39} \\
        \bottomrule
    \end{tabular}
\end{table}
Unfortunately, our findings indicate that these simple,
fixed templates do not yield substantial performance improvements compared to other baselines.
The scores for "The answer is" and "I have found the answer!" are significantly lower than those of our LAQ++ method and even SnapKV.
This suggests that the performance gains from LAQ++ are likely not due to simple semantic guidance provided by the text,
but rather from the pseudo vector information that is more effectively aligned with the output distribution.
Consequently, constructing the Q-Cache through a lookahead mechanism remains a crucial and necessary component of our approach.

\section{RULER Task Evaluation}
Beyond the standard LongBench and "needle in a haystack" benchmarks,
we have included a more challenging evaluation using the RULER task to assess the long-context capabilities of our model.
The experimental setup is consistent with the methodology described in Section \ref{sec:setup}.
The following Table \ref{tab:ruler_task} presents the experimental results for the Llama 3.1-8B-Instruct model:
\begin{table}[t]
\centering
\small
\caption{RULER Task Performance on Llama 3.1-8B-Instruct}
\label{tab:ruler_task}
\begin{tabular}{lcc}
\toprule
\textbf{Method} & \textbf{Budget} & \textbf{Avg.} \\
\midrule
FullKV & 32K & 89.28 \\
\midrule
StreamingLLM & 128 & 12.19 \\
H2O & 128 & 14.40 \\
SnapKV & 128 & 36.87 \\
LAQ++ & 128 & \textbf{51.57} \\
\midrule
StreamingLLM & 256 & 13.31 \\
H2O & 256 & 25.25 \\
SnapKV & 256 & 51.58 \\
LAQ++ & 256 & \textbf{62.81} \\
\midrule
StreamingLLM & 512 & 14.83 \\
H2O & 512 & 36.44 \\
SnapKV & 512 & 60.89 \\
LAQ++ & 512 & \textbf{69.83} \\
\bottomrule
\end{tabular}
\end{table}

As demonstrated by the results,
the RULER task reveals a more significant performance differentiation among the various methods, particularly with a low KV-Cache budget. The proposed method, LAQ++, shows a nearly 10-point improvement over SnapKV when evaluated on this task.

\section{Full Evaluation on LongBench}
\label{app:res}
We present additional experimental results, including those for Qwen2.5-7B-Instruct.
The evaluation is conducted with a context length of 32K for Mistral and Qwen, 8K for Llama3, and 100K for Llama3.1.

\begin{table*}[!t]
\fontsize{20}{24}\selectfont
\setlength{\tabcolsep}{5pt}
\centering
\caption{Performance comparison of different methods across various LLMs on LongBench. The brackets in LAQ denote the Q-Cache length.}
\label{tab:longbenchMistral}
\begin{threeparttable}
\scalebox{0.3}{
\begin{tabular}{llccccccccccccccccc}
\specialrule{1pt}{0pt}{2pt}
&\multirow{4}{*}{\textbf{~~~LLMs}} & \multicolumn{3}{c}{\textbf{Single-Document QA}} & \multicolumn{3}{c}{\textbf{Multi-Document QA}}& \multicolumn{3}{c}{\textbf{Summarization}}& \multicolumn{3}{c}{\textbf{Few-shot Learning}}& \multicolumn{2}{c}{\textbf{Synthetic}} & \multicolumn{2}{c}{\textbf{Code}}&\multirow{4}{*}{\textbf{~~~Avg.}} \\
\cmidrule(lr){3-5}\cmidrule(lr){6-8}\cmidrule(lr){9-11}\cmidrule(lr){12-14}\cmidrule(lr){15-16}\cmidrule(lr){17-18}
&& \rotatebox[origin=c]{30}{\textbf{NrtvQA}} & \rotatebox[origin=c]{30}{\textbf{Qasper}} & \rotatebox[origin=c]{30}{\textbf{MF-en}} & \rotatebox[origin=c]{30}{\textbf{HotpotQA}} & \rotatebox[origin=c]{30}{\textbf{2WikiMQA}} & \rotatebox[origin=c]{30}{\textbf{Musique}} & \rotatebox[origin=c]{30}{\textbf{GovReport}} & \rotatebox[origin=c]{30}{\textbf{QMSum}} & \rotatebox[origin=c]{30}{\textbf{MultiNews}} & \rotatebox[origin=c]{30}{\textbf{TREC}} & \rotatebox[origin=c]{30}{\textbf{TriviaQA}} & \rotatebox[origin=c]{30}{\textbf{SAMSum}} & \rotatebox[origin=c]{30}{\textbf{PCount}} & \rotatebox[origin=c]{30}{\textbf{PRe}} & \rotatebox[origin=c]{30}{\textbf{Lcc}} & \rotatebox[origin=c]{30}{\textbf{RB-P}} \\

\specialrule{1pt}{2pt}{2pt}

\multirow{28}{*}{\rotatebox[origin=c]{90}{\fontsize{18}{100}\selectfont \textbf{Mistral-7B-v0.2-Instruct}}}

&~~~Full KV & 26.90 & 33.14 & 49.26 & 42.77 & 27.35 & 18.77 & 32.95 & 24.21 & 27.13 & 71.00 & 86.23 & 42.72 & 2.75 & 86.98 & 57.12 & 54.51 & 42.74 \\
\cline{2-19}

& \multicolumn{18}{c}{\cellcolor{lightgray!25} 
  \textbf{KV Cache Size = 128}
} \\

&~~~StreamingLLM & 17.12 & 13.43 & 27.31 & 29.54 & 21.91 & 11.94 & 15.61 & 19.18 &17.72 & 44.00 & 79.92 & 37.37 & 3.50 & 23.77 & 51.44 & 45.95 & 28.73 \\

&~~~H2O & 21.62 & 21.34 & 38.61 & 30.46 & 20.38 & 12.20 & 20.59 & 22.51 & 22.03 & 39.00 & 82.33 & 40.64 & 3.06 & \textbf{80.56} & 51.17 & 48.46 & 34.69 \\

&~~~SnapKV & 19.71 & 21.13 & 42.75 & 36.45 & 22.36 & 15.76 & 19.05 & 21.81 & 21.36 & 47.50 & 84.15 & 40.29 & 2.41 & 68.26 & 52.26 & 48.75 & 35.25 \\

&~~~PyramidKV & 21.98 & 22.78 & 43.78 & 32.30 & 22.31 & 15.81 & 20.41 & 21.82 & 21.23 & 66.00 & 83.51 & 39.83 & 2.99 & 65.81 & 51.61 & 46.42 & 36.16 \\

&~~~\textbf{LAQ(2)} & 24.12 & 24.60 & 43.96 & \textbf{40.84} & 25.94 & 17.37 & 21.51 & 21.82 & \textbf{22.81} & \textbf{70.50} & 86.22 & 40.27 & 2.69 & 66.00 & 51.78 & 47.44 & 37.99 \\

&~~~\textbf{LAQ(8)} & 24.94 & \textbf{27.77} & 45.43 & 40.35 & 25.25 & 17.91 & \textbf{22.06} & 22.68 & 22.50 & 69.50 & 86.36 & 41.16 & 1.49 & 76.85 & 53.31 & \textbf{51.02} & 39.29 \\

&~~~\textbf{LAQ(2)++} & 24.40 & 25.32 & 46.06 & 39.37 & \textbf{26.11} & 17.05 & 21.05 & 22.56 & 21.80 & \textbf{70.50} & \textbf{86.39} & 41.19 & 3.33 & 66.46 & 52.33 & 49.03 & 38.31 \\

&~~~\textbf{LAQ(8)++} &\textbf{25.62} & 27.21 & \textbf{46.16} & 40.60 & 25.93 & \textbf{18.44} & 21.60 & \textbf{23.07} & 22.42 & \textbf{70.00} & 86.18 & \textbf{42.03} & \textbf{3.51} & 74.81 & \textbf{54.68} & 50.92 & \textbf{39.57} \\

\cline{2-19}
& \multicolumn{18}{c}{\cellcolor{lightgray!25} 
  \textbf{KV Cache Size = 256}
} \\

&~~~StreamingLLM & 19.22 & 15.08 & 28.01 & 30.40 & 22.02 & 11.21 & 18.09 & 19.59 & 20.09 & 51.00 & 80.71 & 39.89 & 2.96 & 18.40 & 54.30 & 47.38 & 29.90 \\

&~~~H2O & 21.42 & 23.04 & 42.60 & 30.75 & 22.42 & 13.82 & 22.35 & 22.54 & 23.12 & 40.50 & 83.78 & 40.73 & 3.51 & 85.85 & 53.18 & 49.95 & 36.22 \\

&~~~SnapKV & 22.44 & 24.07 & 48.01 & 38.66 & 22.66 & 15.59 & 21.83 & 23.23 & 22.94 & 61.50 & 85.45 & 41.32 & 3.13 & 85.79 & 55.11 & 51.73 & 38.97 \\

&~~~PyramidKV & 21.69 & 25.18 & 47.61 & 38.77 & 26.12 & 15.23 & 22.52 & 22.52 & 22.59 & 68.00 & 84.27 & 42.10 & 3.43 & 76.60 & 53.08 & 48.40 & 38.63 \\

&~~~\textbf{LAQ(2)} & 25.50 & 26.54 & 46.12 & 40.57 & 26.46 & 17.84 & 23.29 & 22.47 & \textbf{23.84} & \textbf{70.50} & 86.03 & 41.98 & \textbf{3.63} & 79.75 & 53.91 & 49.97 & 39.90 \\

&~~~\textbf{LAQ(8)} & 24.68 & \textbf{29.25} & 48.00 & 40.56 & 26.01 & 18.24 & \textbf{24.04} & 22.96 & 23.80 & 70.00 & 85.81 & 42.52 & 2.01 & 82.34 & 54.96 & \textbf{53.00} & 40.51 \\

&~~~\textbf{LAQ(2)++} & \textbf{25.52} & 26.90 & 47.71 & 41.12 & \textbf{27.17} & \textbf{18.93} & 23.02 & 23.24 & 23.25 & \textbf{70.50} & \textbf{86.41} & 41.83 & 3.38 & 79.25 & 54.59 & 50.86 & 40.23 \\

&~~~\textbf{LAQ(8)++} & 25.23 & 29.16 & \textbf{49.24} & \textbf{41.70} & 26.85 & 18.62 & 23.73 & \textbf{23.69} & 23.38 & \textbf{70.50} & 86.24 & \textbf{42.54} & 3.36 & \textbf{86.11} & \textbf{55.59} & 52.49 & \textbf{41.15} \\
\cline{2-19}
& \multicolumn{18}{c}{\cellcolor{lightgray!25} 
  \textbf{KV Cache Size = 512}
} \\

&~~~StreamingLLM & 20.91 & 16.47 & 30.56 & 29.62 & 22.16 & 11.02 & 21.51 & 20.10 & 23.03 & 61.50 & 81.86 & 41.66 & 2.84 & 18.57 & 55.27 & 49.07 & 31.63 \\

&~~~H2O & 21.72 & 25.54 & 44.72 & 32.39 & 23.16 & 14.75 & 23.61 & 23.03 & 24.58 & 42.00 & 85.22 & 41.61 & \textbf{3.42} & 86.45 & 54.82 & 51.11 & 37.38 \\

&~~~SnapKV & 24.14 & 28.11 & 48.78 & 39.49 & 25.09 & 17.40 & 23.67 & 23.18 & 24.60 & 67.00 & 85.88 & 41.39 & 2.78 & \textbf{86.56} & 56.61 & 53.47 & 40.51 \\

&~~~PyramidKV & 22.99 & 28.74 & 48.45 & 39.73 & 25.74 & 16.58 & 24.48 & 23.40 & 24.52 & 70.00 & 85.99 & 42.40 & 3.32 & 81.63 & 55.93 & 52.38 & 40.39 \\

&~~~\textbf{LAQ(2)} & 25.32 & 28.71 & 47.79 & 40.87 & 26.60 & 18.48 & 25.18 & 23.51 & 24.96 & \textbf{71.00} & 86.30 & 42.91 & 2.50 & 83.90 & 55.05 & 51.30 & 40.90 \\

&~~~\textbf{LAQ(8)} & 24.65 & \textbf{31.21} & 49.15 & 39.90 & \textbf{27.18} & 18.38 & 25.55 & 23.91 & 24.87 & \textbf{71.00} & 86.33 & 42.14 & 1.87 & 86.41 & \textbf{56.84} & 53.08 & 41.40 \\

&~~~\textbf{LAQ(2)++} & \textbf{26.24} & 29.51 & 48.32 & 40.67 & 27.11 & 18.98 & 25.05 & 23.48 & 24.72 & \textbf{71.00} & 86.29 & 42.95 & 2.91 & 83.28 & 55.75 & 52.51 & 41.17 \\

&~~~\textbf{LAQ(8)++} & 25.49 & 30.92 & \textbf{49.72} & \textbf{41.50} & 26.84 & \textbf{19.20} & \textbf{25.67} & \textbf{24.04} & \textbf{25.31} & \textbf{71.00} & \textbf{86.43} & \textbf{43.14} & 2.90 & 85.27 & 56.80 & \textbf{53.54} & \textbf{41.70} \\

\midrule

\multirow{28}{*}{\rotatebox[origin=c]{90}{\fontsize{18}{100}\selectfont \textbf{Llama3.1-8B-Instruct}}}

&~~~Full KV & 31.85 & 15.55 & 28.17 & 29.93 & 22.98 & 18.20 & 34.39 & 23.79 & 27.12 & 72.50 & 92.14 & 43.66 & 8.37 & 97.59 & 65.05 & 54.78 & 41.63 \\
\cline{2-19}

& \multicolumn{18}{c}{\cellcolor{lightgray!25} 
  \textbf{KV Cache Size = 128}
} \\

&~~~StreamingLLM & 16.46 & 5.11 & 14.05 & 14.52 & 14.61 & 7.08 & 17.16 & 20.00 & 18.10 & 40.50 & 87.93 & 38.31 & \textbf{11.50} & 95.47 & 58.39 & 47.52 & 31.67 \\

&~~~H2O & 25.06 & 7.09 & 18.58 & 17.86 & 19.88 & 9.14 & 22.28 & 22.68 & 21.55 & 40.00 & 90.89 & 40.78 & 8.30 & 92.96 & 59.15 & 49.36 & 34.10 \\

&~~~SnapKV & 24.65 & 7.29 & 22.01 & 19.11 & 18.85 & 11.07 & 20.48 & 21.62 & 20.16 & 47.50 & 90.24 & 40.47 & 10.75 & 92.51 & 59.99 & 49.08 & 34.74 \\

&~~~PyramidKV & 24.79 & 8.29 & 20.72 & 14.86 & 13.84 & 8.90 & 22.41 & 22.76 & 21.53 & 62.00 & 90.35 & 39.23 & 9.27 & 93.51 & 58.77 & 46.46 & 34.86 \\

&~~~\textbf{LAQ(2)} & 24.74 & \textbf{11.12} & 22.02 & 20.02 & 20.53 & 13.87 & 23.72 & 22.74 & 22.79 & \textbf{72.50} & 91.59 & 41.51 & 8.25 & 95.45 & 60.09 & 47.06 & 37.38 \\

&~~~\textbf{LAQ(8)} & 27.80 & 10.66 & 24.86 & 20.64 & 20.04 & 15.40 & \textbf{24.18} & 23.09 & \textbf{22.88} & \textbf{72.00} & 91.55 & \textbf{43.43} & 9.04 & \textbf{95.85} & 61.12 & 50.33 & 38.30 \\

&~~~\textbf{LAQ(2)++} & \textbf{28.68} & 10.51 & 24.92 & 20.53 & 20.32 & 14.50 & 22.88 & 23.16 & 22.45 & \textbf{72.50} & 91.65 & 41.59 & 8.21 & 91.96 & 61.72 & 49.30 & 37.81 \\

&~~~\textbf{LAQ(8)++} & 28.65 & 10.65 & \textbf{26.04} & \textbf{24.23} & \textbf{21.56} & \textbf{15.67} & 23.50 & \textbf{23.74} & 22.75 & \textbf{72.00} & \textbf{91.95} & 42.19 & 8.37 & 94.81 & \textbf{62.18} & \textbf{50.95} & \textbf{38.70} \\
\cline{2-19}
& \multicolumn{18}{c}{\cellcolor{lightgray!25} 
  \textbf{KV Cache Size = 256}
} \\

&~~~StreamingLLM & 18.76 & 5.47 & 13.88 & 13.81 & 14.32 & 6.73 & 19.94 & 20.19 & 20.85 & 44.50 & 89.01 & 40.51 & \textbf{11.33} & 92.42 & 61.87 & 49.33 & 32.68 \\

&~~~H2O & 26.01 & 8.42 & 19.69 & 17.28 & 18.21 & 9.91 & 23.64 & 22.89 & 23.20 & 41.50 & 91.29 & 41.60 & 8.00 & 94.31 & 60.79 & 50.31 & 34.82 \\

&~~~SnapKV & 28.05 & 9.83 & 22.71 & 21.48 & 19.36 & 10.96 & 22.86 & 22.75 & 22.98 & 58.00 & 92.28 & 40.87 & 8.10 & 95.30 & 63.64 & 51.35 & 36.91 \\

&~~~PyramidKV & 26.40 & 10.08 & 22.46 & 15.20 & 16.38 & 8.60 & 23.86 & 22.93 & 23.17 & 69.00 & 90.99 & 40.60 & 8.42 & 93.74 & 60.59 & 48.11 & 36.28 \\

&~~~\textbf{LAQ(2)} & 28.23 & 11.51 & 24.66 & 21.57 & 22.81 & 15.15 & 24.81 & 23.18 & 23.83 & \textbf{72.50} & 92.25 & 43.13 & 8.52 & 94.30 & 62.01 & 49.93 & 38.65 \\

&~~~\textbf{LAQ(8)} & 28.86 & 12.40 & 26.44 & 21.80 & 20.91 & 15.77 & \textbf{25.83} & 23.30 & \textbf{24.26} & \textbf{72.50} & \textbf{93.08} & 42.57 & 7.53 & \textbf{95.80} & 63.51 & 51.09 & 39.10 \\

&~~~\textbf{LAQ(2)++} & \textbf{30.49} & 11.95 & 25.79 & 22.99 & 21.36 & 15.35 & 24.40 & 23.38 & 23.53 & \textbf{72.50} & 91.97 & 42.45 & 7.50 & 94.49 & 63.65 & 50.55 & 38.90 \\

&~~~\textbf{LAQ(8)++} & 30.25 & \textbf{12.43} & \textbf{26.63} & \textbf{25.77} & \textbf{22.83} & \textbf{18.45} & 25.07 & \textbf{23.67} & 23.75 & 72.00 & 91.97 & \textbf{43.17} & 6.93 & 94.59 & \textbf{64.07} & \textbf{53.38} & \textbf{39.69} \\
\cline{2-19}
& \multicolumn{18}{c}{\cellcolor{lightgray!25} 
  \textbf{KV Cache Size = 512}
} \\

&~~~StreamingLLM & 20.86 & 6.62 & 14.97 & 14.74 & 13.87 & 6.79 & 22.78 & 20.84 & 23.93 & 57.00 & 89.57 & 41.10 & \textbf{11.75} & 93.73 & 63.37 & 51.79 & 34.61 \\

&~~~H2O & 25.44 & 8.35 & 20.97 & 20.08 & 19.23 & 9.51 & 24.44 & 23.50 & 24.35 & 44.00 & 92.10 & 41.16 & 7.43 & 96.41 & 62.73 & 51.77 & 35.72 \\

&~~~SnapKV & 30.34 & 10.75 & 23.54 & 24.65 & 21.55 & 12.98 & 24.82 & 23.15 & 24.61 & 68.00 & \textbf{92.33} & 42.16 & 7.83 & 96.86 & 64.74 & \textbf{53.60} & 38.87 \\

&~~~PyramidKV & 28.38 & 11.59 & 25.02 & 20.06 & 18.80 & 10.64 & 25.73 & 24.03 & 25.01 & 70.00 & 92.22 & 41.73 & 8.47 & 96.42 & 63.44 & 51.02 & 38.29 \\

&~~~\textbf{LAQ(2)} & 30.65 & 12.95 & 26.69 & 25.29 & 22.31 & 16.57 & 26.51 & 22.99 & 25.27 & \textbf{72.50} & 92.25 & 42.88 & 8.11 & 96.66 & 64.09 & 51.56 & 39.83 \\

&~~~\textbf{LAQ(8)} & \textbf{30.89} & \textbf{14.04} & 25.86 & 26.00 & 23.19 & 17.73 & \textbf{27.07} & 24.01 & \textbf{25.29} & \textbf{72.50} & 92.25 & \textbf{43.13} & 6.96 & \textbf{96.97} & 64.87 & 52.58 & 40.21 \\

&~~~\textbf{LAQ(2)++} & 30.85 & 12.95 & 26.03 & 27.31 & 21.66 & 17.73 & 26.40 & 23.51 & 24.83 & \textbf{72.50} & 91.97 & 43.01 & 7.90 & 96.79 & 64.17 & 52.84 & 40.03 \\

&~~~\textbf{LAQ(8)++} & 29.64 & 13.22 & \textbf{26.79} & \textbf{27.58} & \textbf{23.49} & \textbf{18.63} & 26.94 & \textbf{24.04} & 25.21 & \textbf{72.50} & 92.25 & 42.83 & 8.43 & 96.25 & \textbf{65.00} & 53.37 & \textbf{40.39} \\

\specialrule{1pt}{2pt}{0pt}
\end{tabular}
}
\end{threeparttable}\vspace{-10pt}
\end{table*}

\begin{table*}[!t]

\fontsize{20}{24}\selectfont
\setlength{\tabcolsep}{5pt}
\centering
\caption{Performance comparison of different methods across various LLMs on LongBench. The brackets in LAQ denote the Q-Cache length.}\label{tab:all_longbench}
\begin{threeparttable}
\scalebox{0.3}{
\begin{tabular}{llccccccccccccccccc}
\specialrule{1pt}{0pt}{2pt}
&\multirow{4}{*}{\textbf{~~~LLMs}} & \multicolumn{3}{c}{\textbf{Single-Document QA}} & \multicolumn{3}{c}{\textbf{Multi-Document QA}}& \multicolumn{3}{c}{\textbf{Summarization}}& \multicolumn{3}{c}{\textbf{Few-shot Learning}}& \multicolumn{2}{c}{\textbf{Synthetic}} & \multicolumn{2}{c}{\textbf{Code}}&\multirow{4}{*}{\textbf{~~~Avg.}} \\
\cmidrule(lr){3-5}\cmidrule(lr){6-8}\cmidrule(lr){9-11}\cmidrule(lr){12-14}\cmidrule(lr){15-16}\cmidrule(lr){17-18}
&& \rotatebox[origin=c]{30}{\textbf{NrtvQA}} & \rotatebox[origin=c]{30}{\textbf{Qasper}} & \rotatebox[origin=c]{30}{\textbf{MF-en}} & \rotatebox[origin=c]{30}{\textbf{HotpotQA}} & \rotatebox[origin=c]{30}{\textbf{2WikiMQA}} & \rotatebox[origin=c]{30}{\textbf{Musique}} & \rotatebox[origin=c]{30}{\textbf{GovReport}} & \rotatebox[origin=c]{30}{\textbf{QMSum}} & \rotatebox[origin=c]{30}{\textbf{MultiNews}} & \rotatebox[origin=c]{30}{\textbf{TREC}} & \rotatebox[origin=c]{30}{\textbf{TriviaQA}} & \rotatebox[origin=c]{30}{\textbf{SAMSum}} & \rotatebox[origin=c]{30}{\textbf{PCount}} & \rotatebox[origin=c]{30}{\textbf{PRe}} & \rotatebox[origin=c]{30}{\textbf{Lcc}} & \rotatebox[origin=c]{30}{\textbf{RB-P}} \\

\specialrule{1pt}{2pt}{2pt}
\multirow{28}{*}{\rotatebox[origin=c]{90}{\fontsize{18}{100}\selectfont \textbf{Qwen2.5-7B-Instruct}}}

&~~~Full KV & 17.07 & 43.76 & 52.61 & 57.70 & 47.13 & 29.85 & 32.00 & 23.61 & 23.95 & 39.50 & 87.65 & 40.08 & 8.50 & 100.00 & 6.62 & 9.79 & 38.74 \\
\cline{2-19}

& \multicolumn{18}{c}{\cellcolor{lightgray!25} 
  \textbf{KV Cache Size = 128}
} \\

&~~~StreamingLLM & 11.77 & 23.65 & 26.44 & 40.90 & 37.09 & 16.42 & 16.12 & 18.03 & 14.69 & 12.00 & 76.26 & 35.50 & \textbf{8.50} & 24.50 & \textbf{10.26} & \textbf{13.44} & 24.10 \\

&~~~H2O & 16.61 & 29.58 & 37.25 & 51.04 & 42.14 & 20.82 & 20.33 & \textbf{22.05} & 18.07 & 18.50 & 80.85 & \textbf{39.04} & \textbf{8.50} & \textbf{98.00} & 6.29 & 9.77 & 32.43 \\

&~~~SnapKV & 16.23 & 32.72 & 45.14 & 52.95 & 44.10 & 24.16 & 19.27 & 20.81 & 17.27 & 21.50 & 85.83 & 37.75 & \textbf{8.50} & 95.00 & 7.26 & 11.28 & 33.74 \\

&~~~PyramidKV & 15.40 & 30.67 & 44.89 & 51.54 & 41.22 & 25.62 & 19.51 & 19.71 & 16.69 & 25.75 & 84.22 & 34.08 & \textbf{8.50} & 96.00 & 5.29 & 6.32 & 32.84 \\

&~~~\textbf{LAQ(2)} & 15.32 & 35.05 & 46.87 & 54.16 & 42.66 & 25.82 & 21.52 & 20.90 & 18.48 & 33.50 & 87.52 & 37.58 & \textbf{8.50} & 97.00 & 4.90 & 7.25 & 34.81 \\

&~~~\textbf{LAQ(8)} & 17.98 & 39.17 & 48.85 & 56.76 & 44.96 & \textbf{28.96} & \textbf{22.39} & 20.66 & \textbf{18.77} & 35.00 & 87.78 & 37.75 & \textbf{8.50} & \textbf{98.00} & 5.82 & 9.35 & 36.29 \\

&~~~\textbf{LAQ(2)++} & \textbf{18.21} & 36.81 & 48.80 & 55.97 & 44.96 & 28.91 & 21.24 & 21.22 & 18.09 & 33.00 & \textbf{87.96} & 36.52 & \textbf{8.50} & 96.00 & 5.35 & 7.58 & 35.57 \\

&~~~\textbf{LAQ(8)++} & 18.14 & \textbf{39.40} & \textbf{50.26} & \textbf{56.83} & \textbf{46.07} & 28.12 & 21.74 & 21.80 & 18.65 & \textbf{35.50} & 87.56 & 37.06 & \textbf{8.50} & \textbf{98.00} & 6.13 & 9.96 & \textbf{36.48} \\

\cline{2-19}
& \multicolumn{18}{c}{\cellcolor{lightgray!25} 
  \textbf{KV Cache Size = 256}
} \\

&~~~StreamingLLM & 11.50 & 24.44 & 26.81 & 41.85 & 37.25 & 17.39 & 18.90 & 18.12 & 17.11 & 17.50 & 80.66 & 37.47 & \textbf{9.00} & 24.50 & \textbf{8.42} & \textbf{13.54} & 25.28 \\

&~~~H2O & 16.05 & 32.25 & 43.63 & 52.73 & 43.40 & 23.98 & 22.66 & 22.30 & 19.60 & 19.50 & 83.55 & 39.24 & 8.50 & \textbf{99.50} & 6.71 & 9.29 & 33.93 \\

&~~~SnapKV & 16.25 & 36.47 & 50.35 & 55.44 & 44.37 & 26.62 & 21.96 & 21.76 & 19.35 & 31.00 & 86.89 & 38.47 & 8.50 & 98.00 & 6.52 & 9.98 & 35.75 \\

&~~~PyramidKV & 16.81 & 34.46 & 47.17 & 54.72 & 42.74 & 25.86 & 21.45 & 20.39 & 17.61 & 33.75 & 85.76 & 36.93 & 8.50 & 98.50 & 5.15 & 7.21 & 34.81 \\

&~~~\textbf{LAQ(2)} & 16.31 & 39.64 & 47.04 & 57.01 & 44.78 & 27.35 & 23.86 & 21.59 & 19.69 & 36.00 & 87.81 & 38.98 & 8.50 & 98.50 & 5.35 & 7.51 & 36.25 \\

&~~~\textbf{LAQ(8)} & \textbf{17.62} & \textbf{41.16} & 50.84 & 57.16 & \textbf{46.48} & \textbf{29.37} & \textbf{24.23} & 22.17 & \textbf{20.30} & 38.50 & \textbf{88.10} & \textbf{39.12} & 8.50 & 99.00 & 6.14 & 9.83 & \textbf{37.41} \\

&~~~\textbf{LAQ(2)++} & 16.79 & 38.96 & 49.78 & 57.13 & 45.59 & 27.95 & 23.27 & 21.33 & 19.92 & \textbf{39.00} & 87.51 & 37.82 & 8.50 & 98.50 & 5.40 & 8.07 & 36.60 \\

&~~~\textbf{LAQ(8)++} & 16.76 & 41.04 & \textbf{51.23} & \textbf{57.50} & 45.94 & 28.28 & 24.09 & \textbf{22.36} & 20.24 & 38.00 & 87.61 & 38.54 & 8.50 & 98.50 & 6.93 & 9.84 & 37.21 \\
\cline{2-19}
& \multicolumn{18}{c}{\cellcolor{lightgray!25} 
  \textbf{KV Cache Size = 512}
} \\

&~~~StreamingLLM & 12.56 & 27.00 & 29.75 & 42.48 & 36.44 & 16.15 & 22.30 & 18.68 & 20.01 & 24.00 & 83.99 & 38.09 & \textbf{8.50} & 24.00 & \textbf{8.36} & \textbf{12.54} & 26.55 \\

&~~~H2O & 16.77 & 35.32 & 45.82 & 52.30 & 44.39 & 24.29 & 24.20 & 22.29 & 21.45 & 21.00 & 86.29 & 39.16 & \textbf{8.50} & \textbf{100.00} & 6.82 & 9.09 & 34.86 \\

&~~~SnapKV & 17.63 & 40.73 & 50.71 & 55.63 & 45.18 & 27.85 & 24.41 & 22.84 & 21.16 & 38.50 & \textbf{88.12} & 39.21 & \textbf{8.50} & 99.50 & 7.09 & 10.16 & 37.33 \\

&~~~PyramidKV & 17.49 & 37.81 & 50.45 & 56.36 & 44.64 & 27.46 & 23.00 & 20.97 & 19.60 & 35.25 & 87.62 & 37.82 & \textbf{8.50} & 99.00 & 5.62 & 7.94 & 36.22 \\

&~~~\textbf{LAQ(2)} & 16.46 & 41.86 & 49.68 & 57.64 & 46.46 & 27.79 & 25.71 & 21.72 & 21.58 & 38.50 & 87.55 & \textbf{40.02} & \textbf{8.50} & \textbf{100.00} & 5.45 & 8.42 & 37.33 \\

&~~~\textbf{LAQ(8)} & 17.26 & 42.64 & \textbf{51.73} & 57.56 & 46.59 & 29.12 & \textbf{26.57} & 22.67 & 21.86 & \textbf{39.00} & 87.65 & 39.40 & \textbf{8.50} & \textbf{100.00} & 6.52 & 9.91 & \textbf{37.94} \\

&~~~\textbf{LAQ(2)++} & \textbf{17.92} & 41.84 & 50.99 & \textbf{57.76} & 46.74 & 28.72 & 25.30 & 22.05 & 21.63 & 38.75 & 87.65 & 39.13 & \textbf{8.50} & 99.50 & 5.56 & 8.25 & 37.52 \\

&~~~\textbf{LAQ(8)++} & 17.31 & \textbf{43.32} & 51.17 & 57.61 & \textbf{46.81} & \textbf{29.29} & 25.98 & \textbf{22.97} & \textbf{21.96} & \textbf{39.00} & 87.65 & 39.01 & \textbf{8.50} & 99.50 & 6.75 & 10.11 & 37.93 \\

\midrule

\multirow{22}{*}{\rotatebox[origin=c]{90}{\fontsize{18}{100}\selectfont \textbf{Llama3-8B-Instruct}}}

&~~~Full KV & 25.56 & 32.27 & 39.71 & 43.56 & 35.09 & 21.18 & 28.71 & 23.26 & 26.64 & 73.50 & 90.48 & 42.33 & 4.80 & 69.25 & 59.29 & 54.05 & 41.86 \\
\cline{2-19}

& \multicolumn{18}{c}{\cellcolor{lightgray!25} 
  \textbf{KV Cache Size = 128}
} \\

&~~~StreamingLLM & 18.13 & 8.54 & 21.21 & 32.86 & 26.27 & 15.41 & 16.71 & 20.46 & 18.06 & 45.00 & 74.58 & 36.32 & 5.75 & 68.50 & 56.12 & 53.06 & 32.31 \\

&~~~H2O & 22.12 & 13.19 & 29.53 & 37.42 & 32.71 & 18.25 & 20.49 & 22.03 & 21.11 & 38.50 & 87.75 & 39.14 & 5.83 & \textbf{69.50} & 57.01 & 54.74 & 35.58 \\

&~~~SnapKV & 20.96 & 13.63 & 30.75 & 36.65 & 29.24 & 19.12 & 19.07 & 21.67 & 20.19 & 45.00 & 87.82 & 38.01 & 5.13 & 69.35 & 57.51 & 55.31 & 35.59 \\

&~~~PyramidKV & 21.40 & 16.92 & 31.62 & 38.45 & 28.72 & 18.59 & 19.96 & \textbf{22.49} & 20.96 & 66.50 & 89.35 & 38.43 & \textbf{5.92} & 69.00 & 57.86 & 51.80 & 37.37 \\

&~~~\textbf{LAQ(1)++} & \textbf{25.01} & 16.34 & 33.73 & \textbf{42.92} & 35.00 & 19.54 & 20.23 & 22.32 & 21.73 & 73.50 & 90.25 & 39.18 & 5.18 & \textbf{69.50} & 59.05 & 54.61 & 39.26 \\

&~~~\textbf{LAQ(8)++} & 24.75 & \textbf{18.87} & \textbf{34.84} & 41.13 & \textbf{36.50} & \textbf{20.13} & \textbf{21.32} & 22.20 & \textbf{22.32} & \textbf{74.00} & \textbf{90.37} & \textbf{40.38} & 5.39 & \textbf{69.50} & \textbf{60.94} & \textbf{57.64} & \textbf{40.02} \\

\cline{2-19}
& \multicolumn{18}{c}{\cellcolor{lightgray!25} 
  \textbf{KV Cache Size = 256}
} \\

&~~~StreamingLLM & 17.98 & 11.10 & 20.58 & 33.68 & 26.16 & 16.03 & 19.24 & 20.46 & 20.80 & 52.50 & 80.18 & 39.31 & \textbf{5.83} & 68.37 & 58.56 & 54.46 & 34.08 \\

&~~~H2O & 23.82 & 16.61 & 31.66 & 38.64 & 31.72 & 20.05 & 21.28 & 22.22 & 22.19 & 39.00 & 89.22 & 39.52 & 5.57 & 69.50 & 58.01 & 54.28 & 36.46 \\

&~~~SnapKV & 24.35 & 18.32 & 33.83 & 42.23 & 32.89 & 20.73 & 20.74 & 22.05 & 22.54 & 62.00 & 90.14 & 39.51 & 5.75 & \textbf{70.00} & 59.76 & 56.66 & 38.84 \\

&~~~PyramidKV & 23.99 & 20.51 & 36.06 & 42.47 & 31.34 & 20.28 & 21.37 & \textbf{22.69} & 22.79 & 71.00 & \textbf{90.48} & 39.86 & 5.83 & 69.25 & 58.64 & 54.06 & 39.41 \\

&~~~\textbf{LAQ(1)++} & 24.57 & 21.01 & 35.80 & 43.52 & 35.23 & 20.62 & 21.61 & 22.20 & 23.18 & 73.50 & 90.44 & 40.32 & 5.50 & 69.70 & 59.46 & 56.62 & 40.21 \\

&~~~\textbf{LAQ(8)++} & \textbf{24.51} & \textbf{22.56} & \textbf{37.15} & \textbf{44.31} & \textbf{36.79} & \textbf{21.60} & \textbf{22.61} & 22.30 & \textbf{23.47} & \textbf{73.50} & 90.37 & \textbf{41.46} & 5.64 & 69.70 & \textbf{62.04} & \textbf{58.00} & \textbf{41.00} \\

\cline{2-19}
& \multicolumn{18}{c}{\cellcolor{lightgray!25} 
  \textbf{KV Cache Size = 512}
} \\

&~~~StreamingLLM & 20.70 & 12.16 & 22.06 & 35.93 & 26.75 & 15.79 & 21.00 & 20.62 & 23.73 & 62.00 & 83.36 & 39.98 & 5.35 & 67.97 & 60.32 & 55.13 & 35.80 \\

&~~~H2O & 23.52 & 17.89 & 33.52 & 41.71 & 33.56 & 19.27 & 22.17 & 22.64 & 23.83 & 41.00 & 90.46 & 40.20 & 5.87 & 69.50 & 58.14 & 56.01 & 37.46 \\

&~~~SnapKV & 24.85 & 23.49 & 36.53 & 42.96 & 34.93 & 20.28 & 22.40 & 22.66 & 23.80 & 70.50 & 90.52 & 40.39 & 5.81 & \textbf{70.00} & 60.45 & 56.17 & 40.36 \\

&~~~PyramidKV & 24.83 & 23.32 & 35.19 & 43.29 & 31.87 & 20.55 & 23.41 & \textbf{22.80} & 24.29 & 71.50 & 90.61 & 40.81 & \textbf{5.91} & 69.50 & 59.60 & 54.71 & 40.14 \\

&~~~\textbf{LAQ(1)++} & 24.96 & 25.91 & 37.07 & 43.19 & \textbf{37.09} & 21.49 & 23.10 & 22.61 & 24.26 & \textbf{73.50} & \textbf{90.64}& 41.61 & 5.43 & 69.70 & 60.89 & 57.83 & 41.21 \\

&~~~\textbf{LAQ(8)++} & \textbf{25.53} & \textbf{27.59} & \textbf{37.99} & \textbf{43.82} & 36.52 & \textbf{21.55} & \textbf{23.58} & 22.64 & \textbf{24.63} & \textbf{73.50} & \textbf{90.64} & \textbf{42.34} & 5.13 & 69.70 & \textbf{62.77} & \textbf{58.45} & \textbf{41.65} \\

\specialrule{1pt}{2pt}{0pt}
\end{tabular}
}
\end{threeparttable}\vspace{-10pt}
\end{table*}

\section{Evaluation on Needle-in-a-Haystack}
\label{app:needle}

Due to space limitations,
we present the complete results of the needle-in-a-haystack tests in the appendix.
The evaluation covers models from the Mistral, Qwen, and LLaMA families,
along with a range of representative KV cache eviction methods.

\begin{figure*}[t]\footnotesize
  \centering
  \begin{subfigure}{\linewidth}
    \centering
    \includegraphics[width=\linewidth, height=5.2cm]{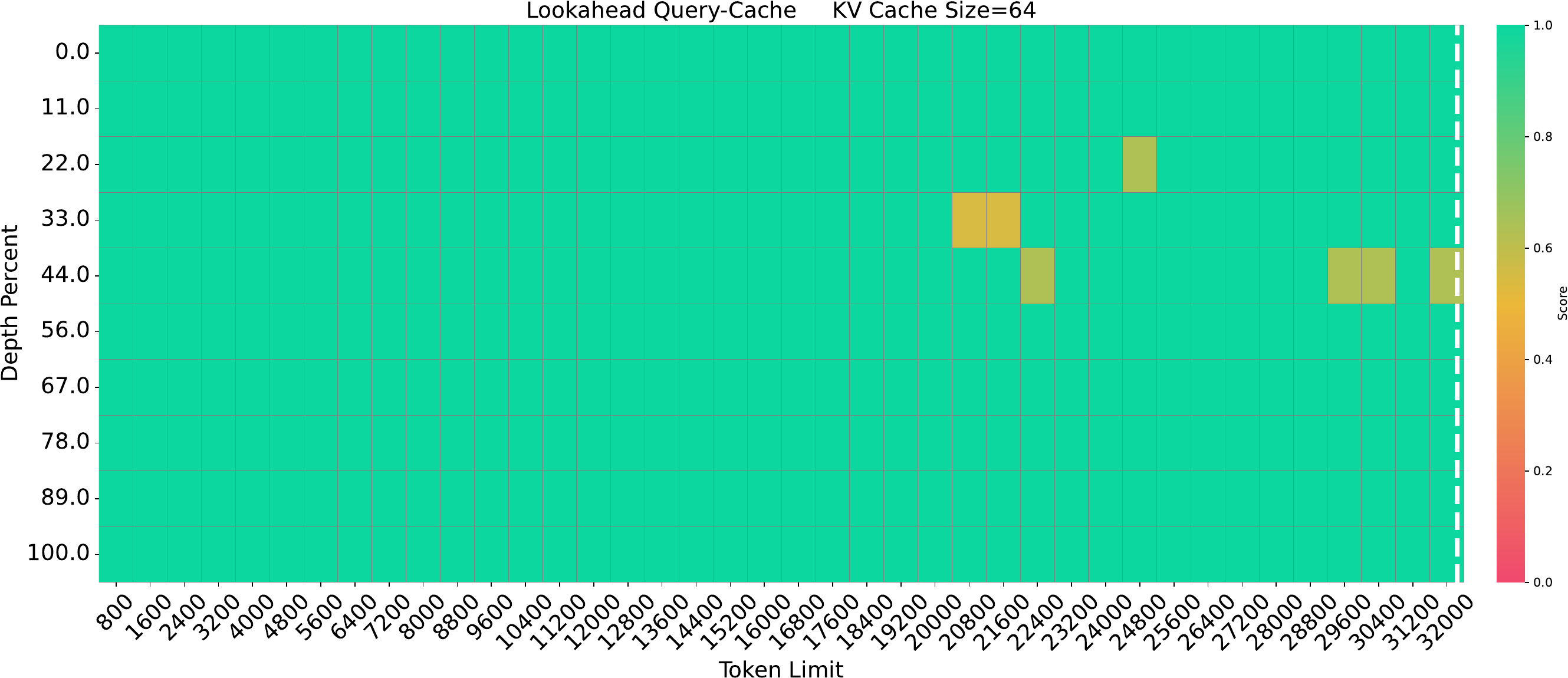}
    \caption{Mistral LAQ++ (Score=99.3)}
  \end{subfigure}
  \hfil
  \begin{subfigure}{\linewidth}
    \centering
    \includegraphics[width=\linewidth, height=5.2cm]{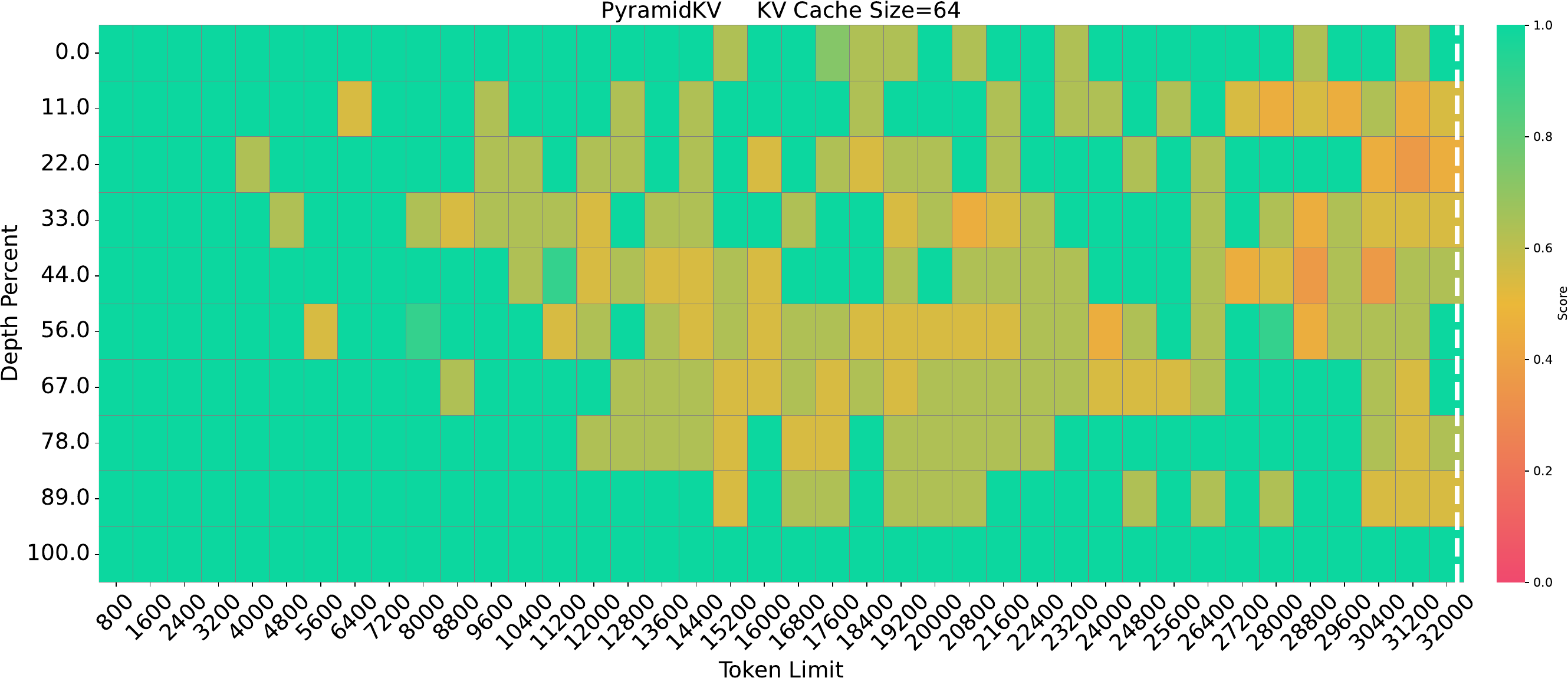}
    \caption{Mistral PyramidKV (Score=84.3)}
  \end{subfigure}
  \hfil
  \begin{subfigure}{\linewidth}
    \centering
    \includegraphics[width=\linewidth, height=5.2cm]{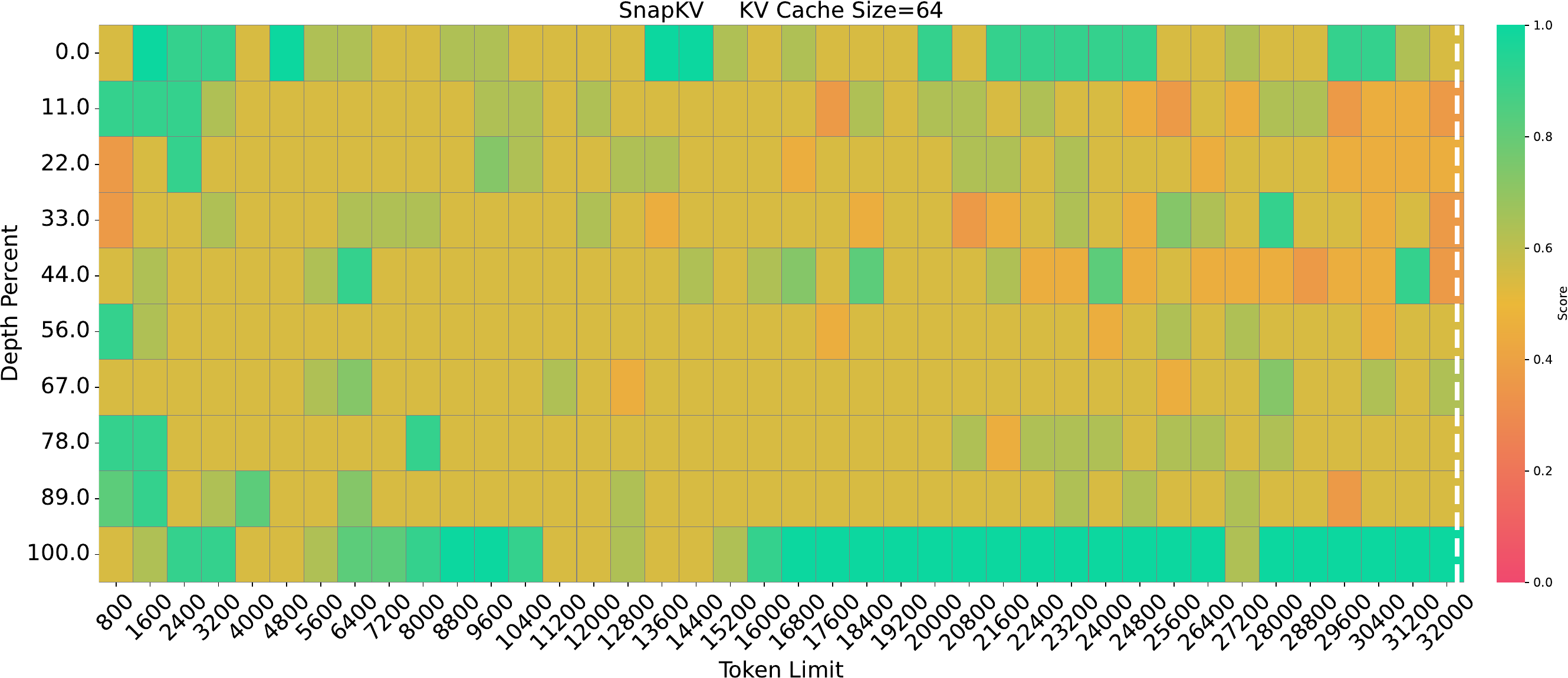}
    \caption{Mistral SnapKV (Score=60.7)}
  \end{subfigure}
  \hfil
  \begin{subfigure}{\linewidth}
    \centering
    \includegraphics[width=\linewidth, height=5.2cm]{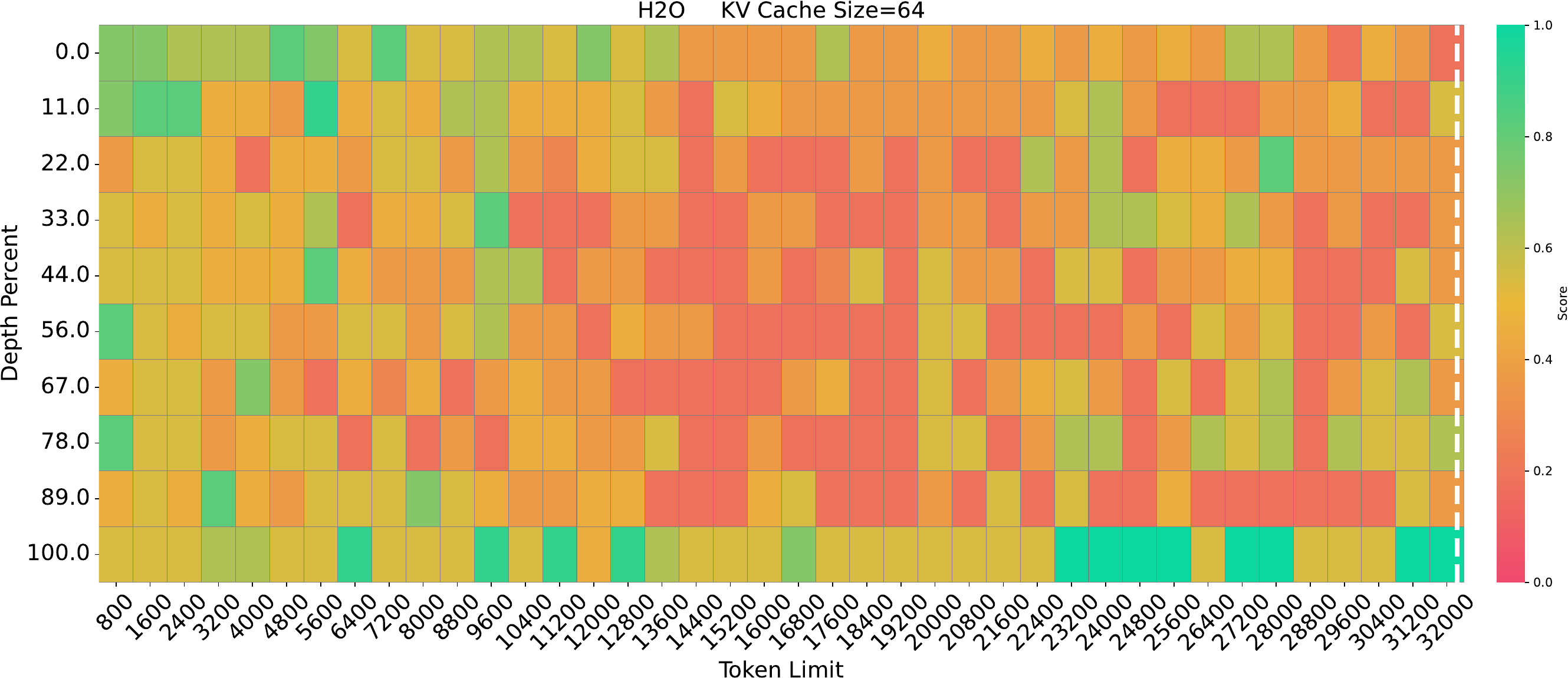}
    \caption{Mistral H2O (Score=43.2)}
  \end{subfigure}
  \caption{The results of different methods of Mistral-7B-v0.2-Instruct on the needle-in-a-haystack with 32k context size under a budget setting of 64.}
\end{figure*}

\begin{figure*}[t]\footnotesize
  \centering
  \begin{subfigure}{\linewidth}
    \centering
    \includegraphics[width=\linewidth, height=5.2cm]{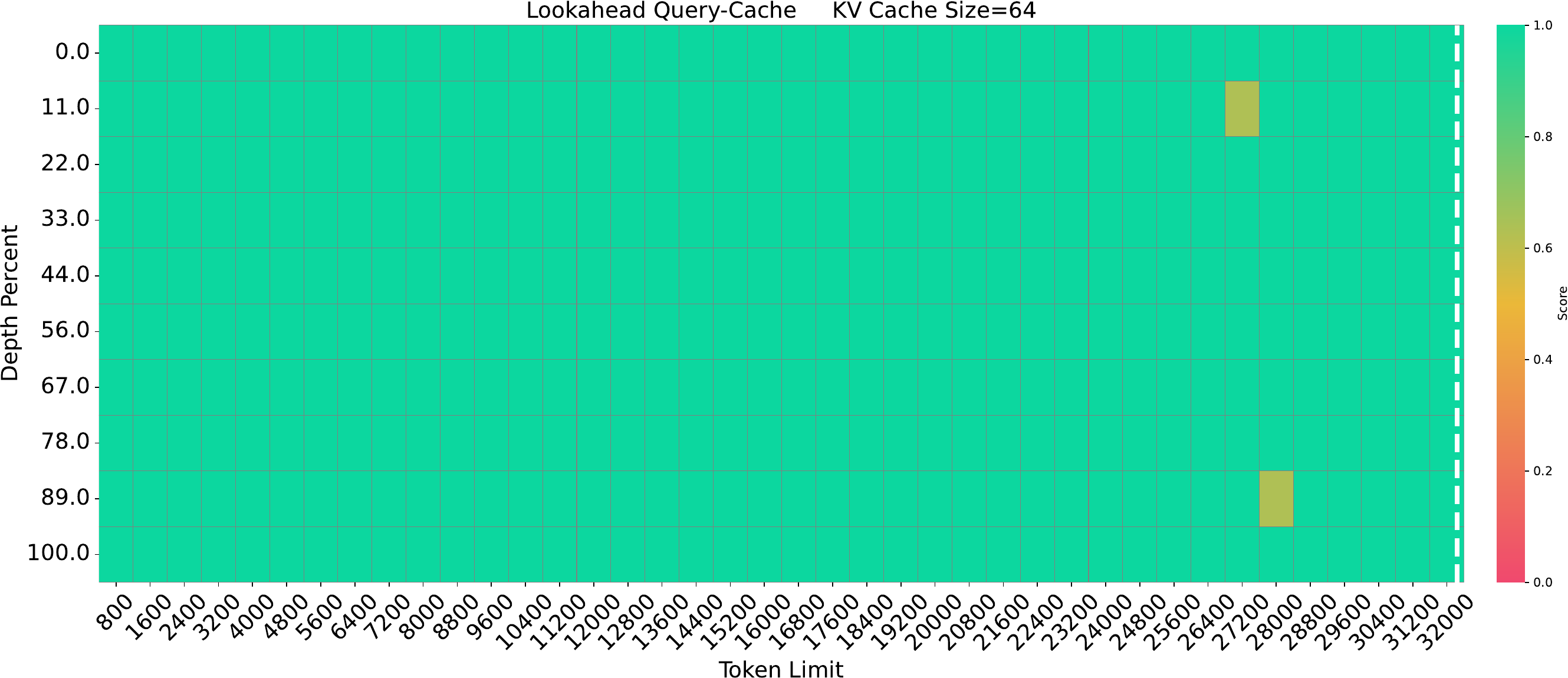}
    \caption{Llama3.1 LAQ++ (Score=99.8)}
  \end{subfigure}
  \hfil
  \begin{subfigure}{\linewidth}
    \centering
    \includegraphics[width=\linewidth, height=5.2cm]{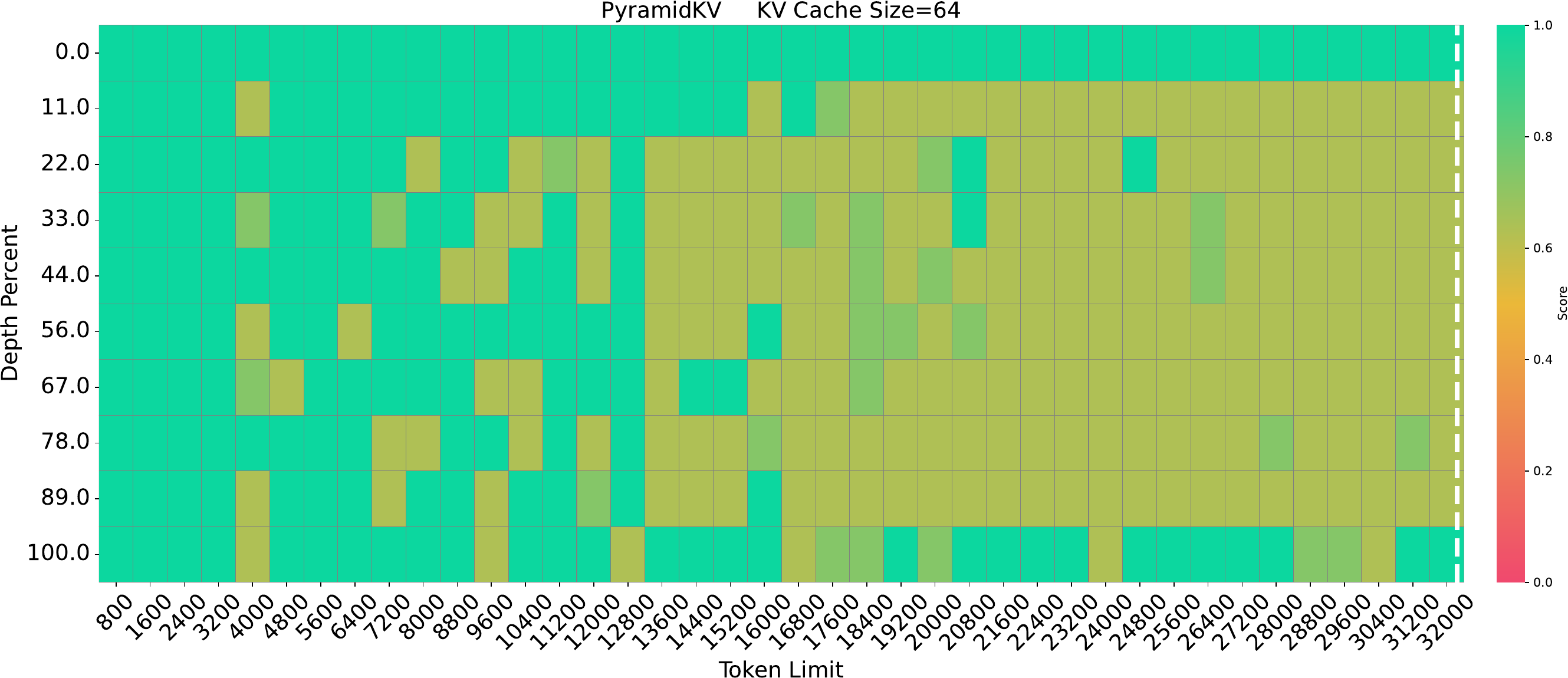}
    \caption{Llama3.1 PyramidKV (Score=80.7)}
  \end{subfigure}
  \hfil
  \begin{subfigure}{\linewidth}
    \centering
    \includegraphics[width=\linewidth, height=5.2cm]{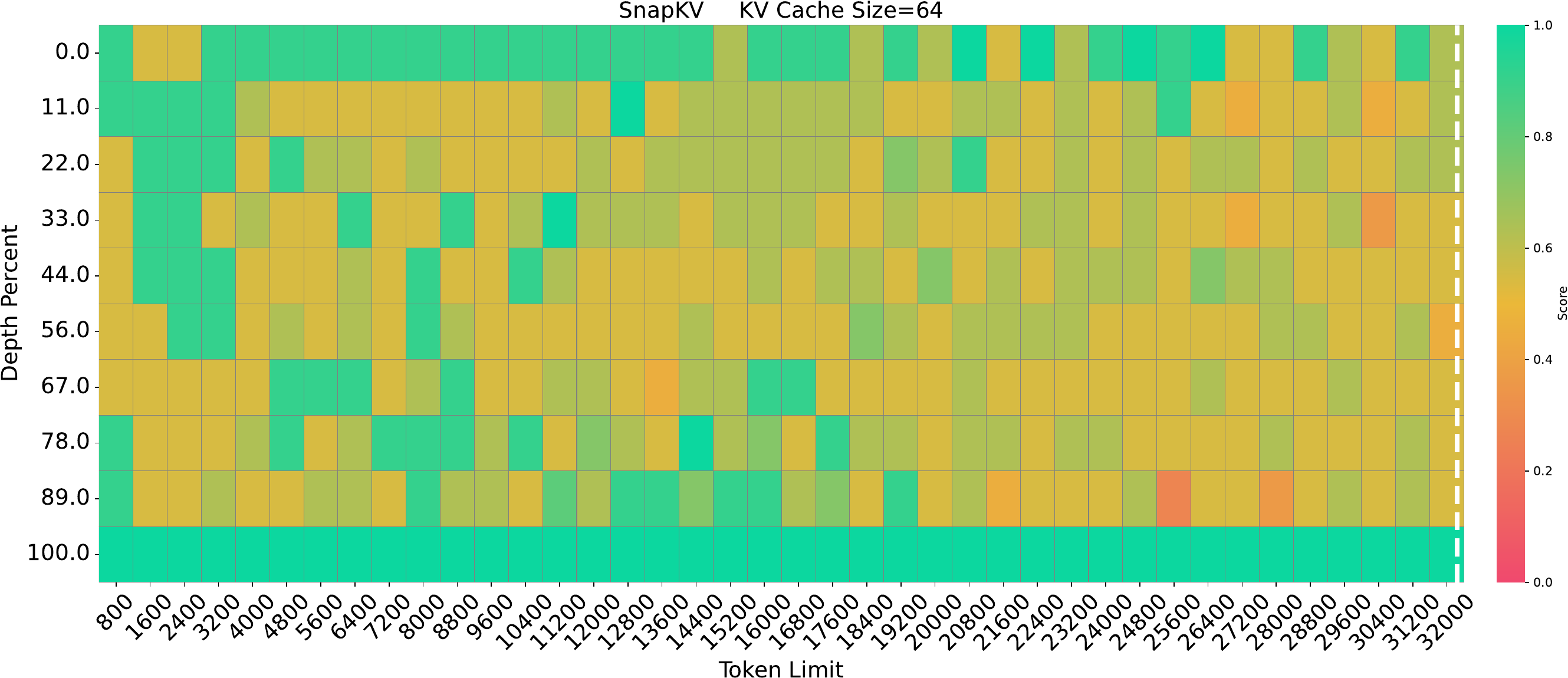}
    \caption{Llama3.1 SnapKV (Score=68.4)}
  \end{subfigure}
  \hfil
  \begin{subfigure}{\linewidth}
    \centering
    \includegraphics[width=\linewidth, height=5.2cm]{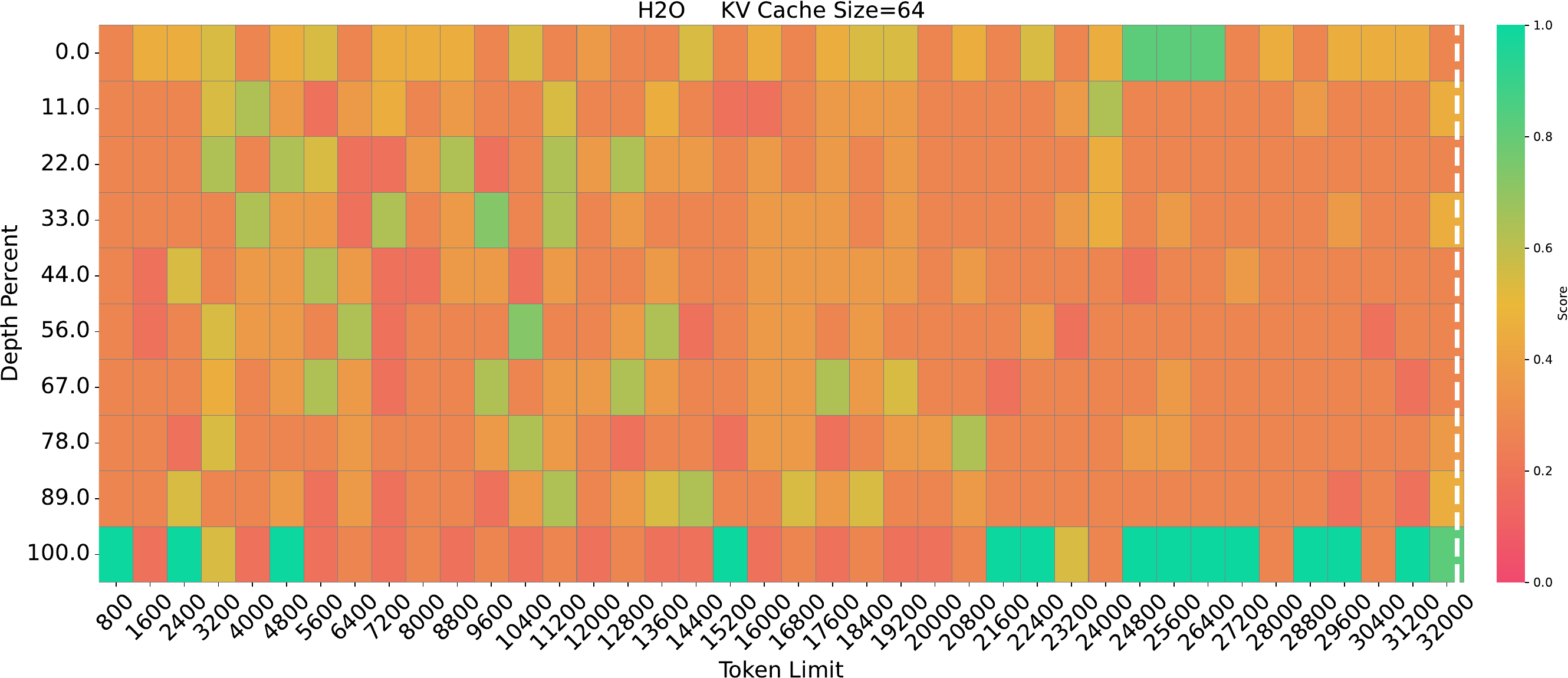}
    \caption{Llama3.1 H2O (Score=35.4)}
  \end{subfigure}
  \caption{The results of different methods of Llama3.1-8B-Instruct on the needle-in-a-haystack with 32k context size under a budget setting of 64.}
\end{figure*}

\begin{figure*}[t]\footnotesize
  \centering
  \begin{subfigure}{\linewidth}
    \centering
    \includegraphics[width=\linewidth, height=5.2cm]{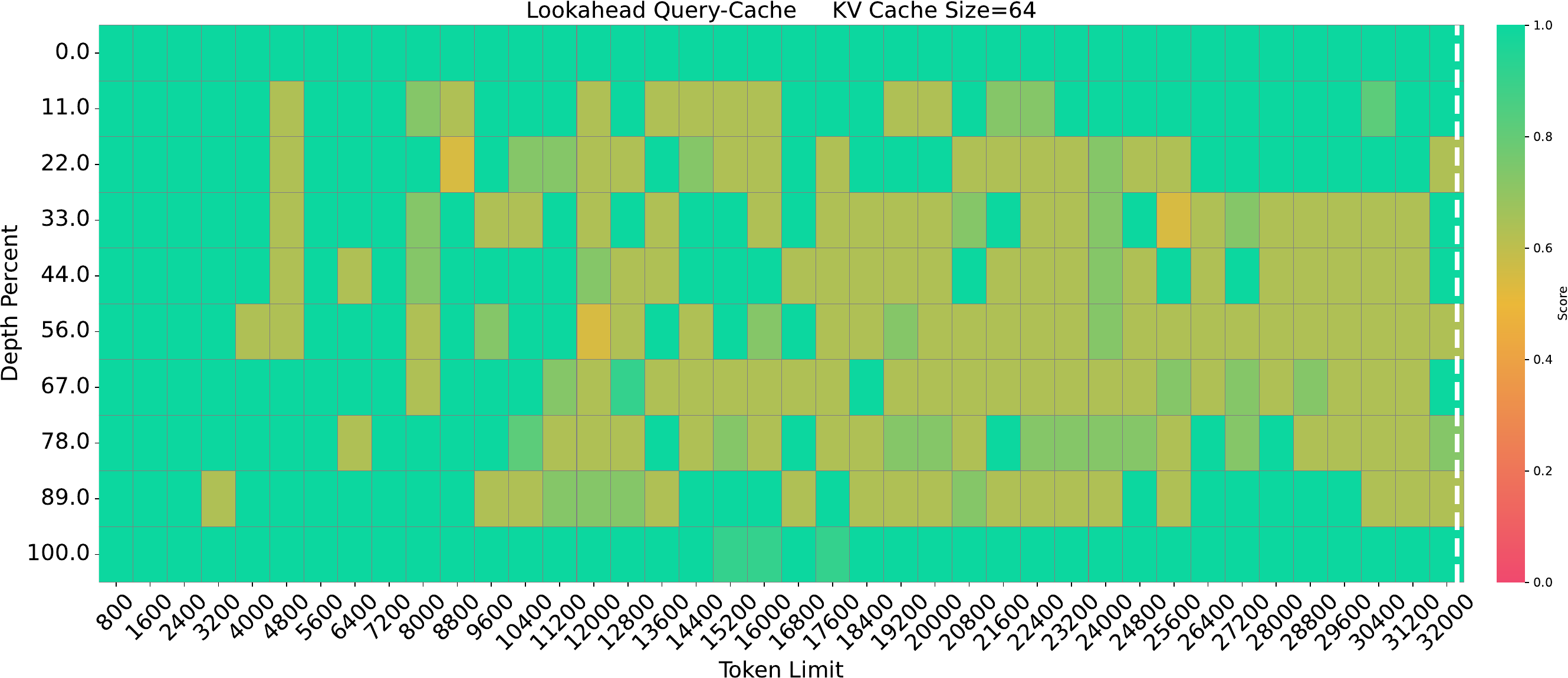}
    \caption{Qwen2.5 LAQ++ (Score=85.1)}
  \end{subfigure}
  \hfil
  \begin{subfigure}{\linewidth}
    \centering
    \includegraphics[width=\linewidth, height=5.2cm]{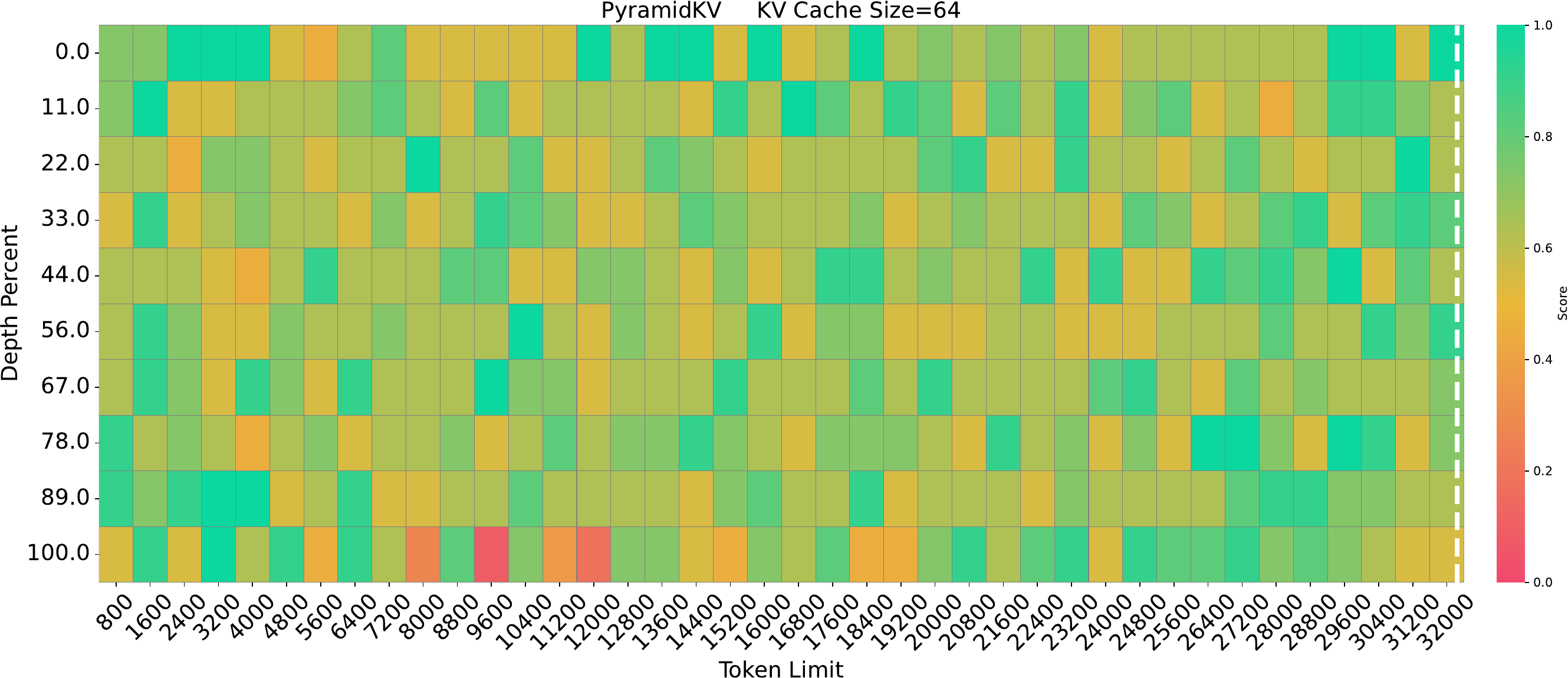}
    \caption{Qwen2.5 PyramidKV (Score=69.3)}
  \end{subfigure}
  \hfil
  \begin{subfigure}{\linewidth}
    \centering
    \includegraphics[width=\linewidth, height=5.2cm]{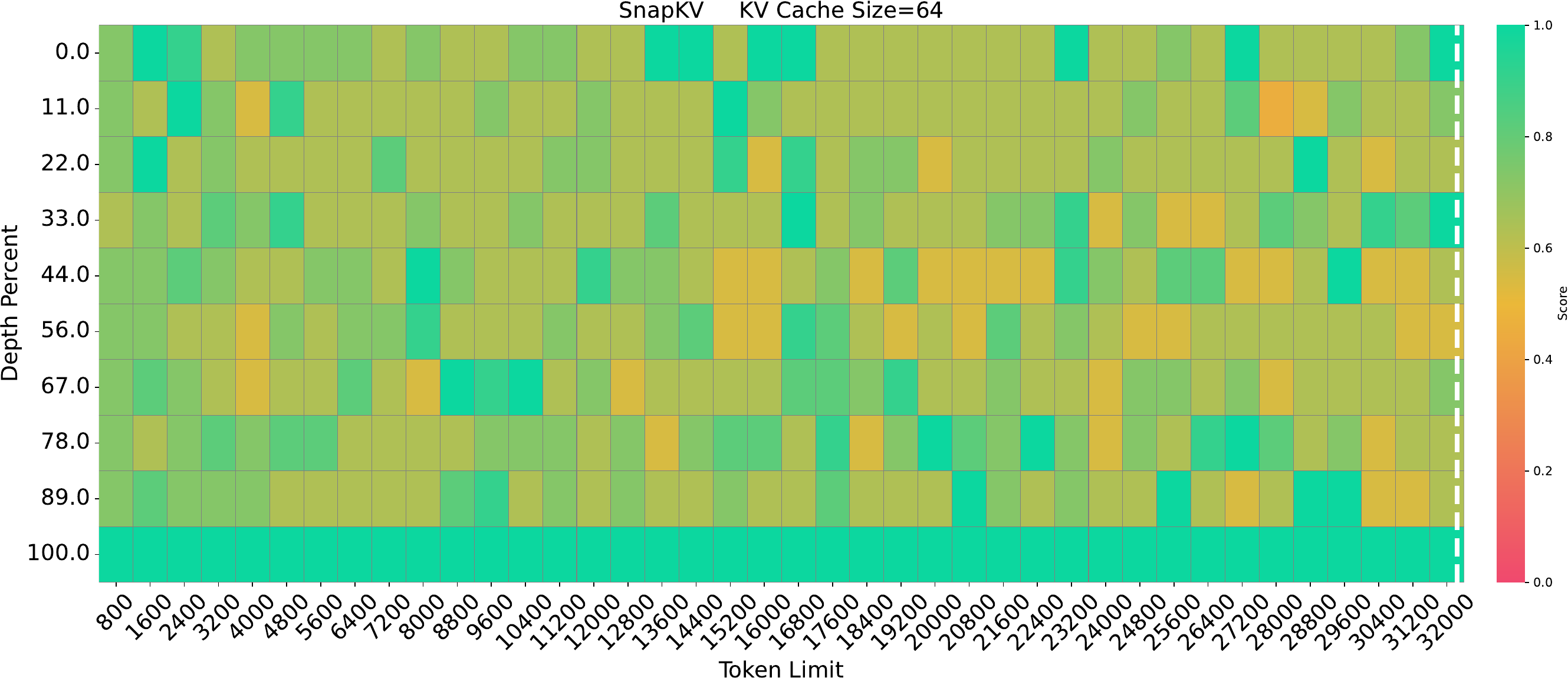}
    \caption{Qwen2.5 SnapKV (Score=72.8)}
  \end{subfigure}
  \hfil
  \begin{subfigure}{\linewidth}
    \centering
    \includegraphics[width=\linewidth, height=5.2cm]{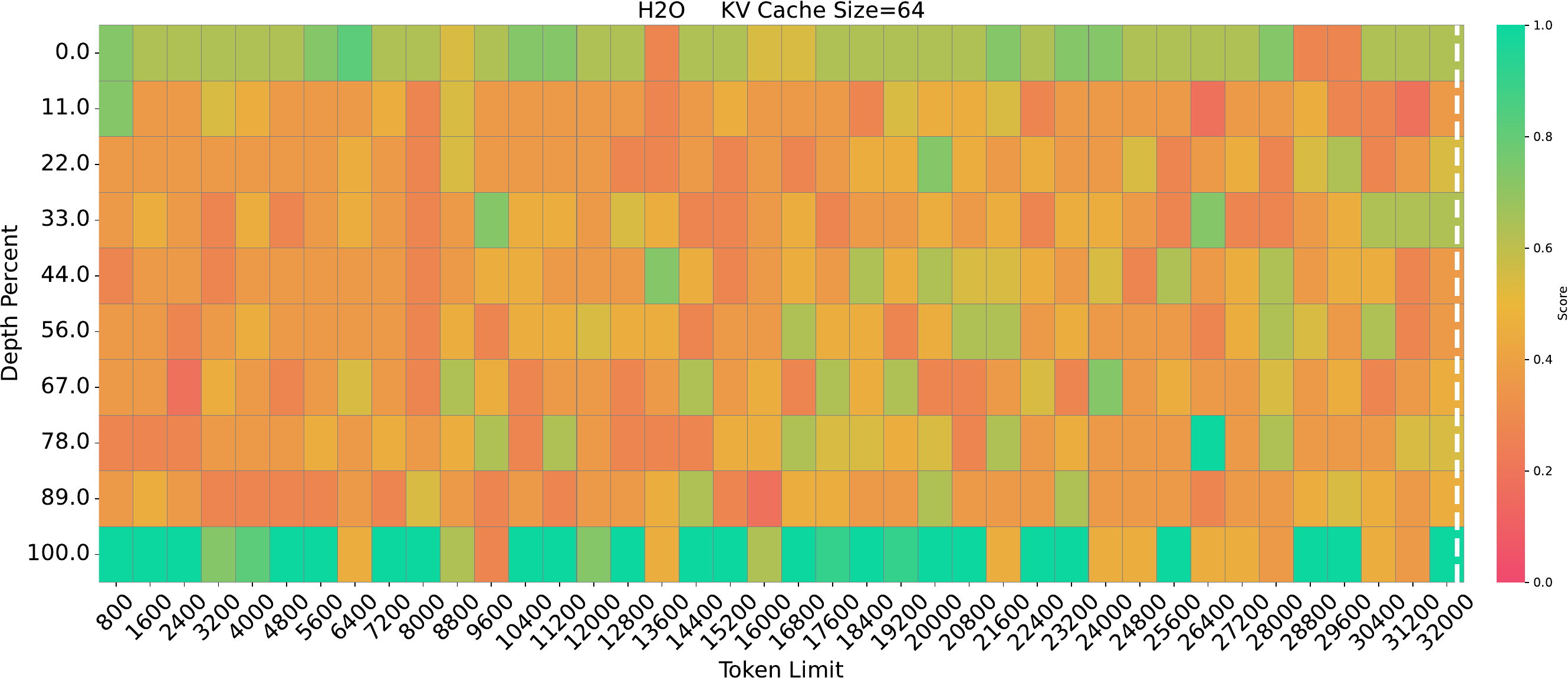}
    \caption{Qwen2.5 H2O (Score=46.7)}
  \end{subfigure}
  \caption{The results of different methods of Qwen2.5-7B-Instruct on the needle-in-a-haystack with 32k context size under a budget setting of 64.}
\end{figure*}

\end{document}